\documentclass[letterpaper, 10 pt, conference]{ieeeconf}
\pdfoutput=1
\usepackage{hyperref}
\usepackage{algorithm}
\usepackage{algpseudocode}
\usepackage{subfig}
\usepackage{graphicx}
\usepackage{algorithmicx}
 
\usepackage{amsthm}
\usepackage{cite}
\usepackage{amsmath,amssymb,amsfonts}
\usepackage{textcomp}
\usepackage{xspace}
\usepackage{makecell}
\usepackage{booktabs,siunitx}
\usepackage{diagbox}
\usepackage{array, multirow}

\newcommand{\gcomb}{\texttt{GComb}\xspace}
\newcommand{\eco}{\texttt{ECO-DQN}\xspace}

\newcommand{\ecosp}{\texttt{ECO-DQN-sparse}\xspace}
\newcommand{\svdqn}{\texttt{S2V-DQN}\xspace}
\newcommand{\lssnn}{\texttt{RELS-DQN}\xspace}

\newcommand{\rels}{\texttt{Robust, Efficient Local-Search}\xspace}
\newcommand{\grls}{\texttt{Greedy-LS}\xspace}

\newcommand{\grev}{\texttt{Greedy-Rev}\xspace}
\newcommand{\sg}{\texttt{Greedy}\xspace}
\newcommand{\relu}{\text{ReLu}\xspace}

\begin{document}

\title{RELS-DQN: A Robust and Efficient Local Search Framework for Combinatorial Optimization}
\author{
\textbf{Yuanhang Shao$^{1, *}~~~$
Tonmoy Dey$^{1, *}~~~$}
\vspace{0.5em} \\
Nikola Vuckovic$^1~~~$
Luke Van Popering$^2~~~$
Alan Kuhnle$^3~~~$
\smallskip 
\vspace{0.25em} \\
$^1$\small{Department of Computer Science, Florida State University}
\\
$^2$\small{Department of Scientific Computing, Florida State University}
\\
$^3$\small{Department of Computer Science, Texas A\&M University}
\\
\small{\{yshao2, tdey, nv21g, lav17b\}@fsu.edu}, \small{kuhnle@tamu.edu}
\vspace{0.25em} \\
$^*$\small{These Two Authors made an equal contribution as Joint First Authors} \vspace{-1.5em}
}

\maketitle

\begin{abstract}
Combinatorial optimization (CO) aims to efficiently find the best solution to NP-hard problems ranging from statistical physics to social media marketing. A wide range of CO applications can benefit from local search methods because they allow reversible action over greedy policies. Deep Q-learning (DQN) using message-passing neural networks (MPNN) has shown promise in replicating the local search behavior and obtaining comparable results to the local search algorithms. However, the over-smoothing and the information loss during the iterations of message passing limit its robustness across applications, and the large message vectors result in memory inefficiency. Our paper introduces \lssnn, a lightweight DQN framework that exhibits the local search behavior while providing practical scalability. Using the \lssnn model trained on one application, it can generalize to various applications by providing solution values higher than or equal to both the local search algorithms and the existing DQN models while remaining efficient in runtime and memory.
\end{abstract}

\section{Introduction}
Combinatorial optimization is a broad and challenging field with real-world applications ranging from traffic routing to recommendation engines. As these problems are often NP-hard \cite{goemans1995improved, dinur2005hardness, hoffman2013traveling, abe2019solving, barrett2020exploratory}, an efficient algorithm to find the best solution in all instances with feasible resources is unlikely to exist. Therefore, researchers have turned to design heuristics \cite{bello2016neural, dai2017learning, barrett2020exploratory, cappart2020combining, yao2021reversible} in addition to approximation algorithms \cite{lee2009non, gupta2010constrained, mirzasoleiman2018streaming, fahrbach2019non, kuhnle2021nearly, chen2021best, dey2022dash} and enumeration \cite{galluccio2001optimization, lazarev2018evaluating}. Among many well-known algorithms, the standard greedy algorithm (\sg) \cite{nemhauser1978analysis} provides the optimal $(1-1/e)$-approximation ratio for monotone submodular instances, but this theoretical guarantee does not hold for non-submodular functions\cite{bian2017guarantees}. The limitation of \sg has led to the development of greedy local search techniques that provide a feasible solution for various applications. These techniques usually allow deletion and exchange operations after the maxima singleton chose greedily, and they have shown promising solution value \cite{yao2021reversible, lee2009non}, but it requires an additional $O(n)$ queries to the objective function $f$. Furthermore, heuristics are usually fast and effective in practice even though it has no theoretical guarantees and requires trial and error over specific problems. In this paper, we focus on heuristically solving the cardinality-constrained maximization problem, which can be described as follows: Given a ground set $\mathcal{N}$, a cardinality parameter $k$, and a set of feasible solutions $\mathcal{S} \subseteq 2^\mathcal{N}$, the goal is to return the solution $V \in \mathcal{S}$ that maximizes the objective function $f: 2^\mathcal{N}\rightarrow \mathbb{R}$ subject to $|V|\leq k$.

\begin{table*}[t]
\centering
\caption{Performance ratio in terms of average solution value of the MaxCut-trained \lssnn to the application-specific trained \eco and approximation algorithms. Across all applications, \lssnn provides a nearly identical solution to \grls with an order-of-magnitude speedup in runtime (Section \ref{efficiency}, and `-' means out of time limit). }\label{tab:summary}
\begin{tabular}{l|ccc|ccc|ccc|ccc}
\hline
\rule{0pt}{15pt}\multirow{2}{*}{Application} & \multicolumn{3}{c|}{\textbf{\Large$\frac{\lssnn}{\sg}$}} & \multicolumn{3}{c|}{\textbf{\Large$\frac{\lssnn}{\grev}$}} & \multicolumn{3}{c|}{\textbf{\Large$\frac{\lssnn}{\grls}$}} & \multicolumn{3}{c}{\textbf{\Large$\frac{\lssnn}{\eco}$}} \\ \cline{2-13} 
                             & ER     & BA     & Large & ER      & BA     & Large  & ER      & BA     & Large  & ER          & BA          & Large       \\ \hline\hline
Maxcut                       & 1.053  & 1.066  & 1.002 & 1.034   & 1.054  & 1.000  & 1.020   & 1.050  & -      & 0.993       & 1.001       & 1.004       \\
MaxCov                       & 1.071  & 1.215  & 1.139 & 1.056   & 1.052  & 1.017  & 1.032   & 1.015  & 0.976  & 1.024       & 1.119       & 1.184       \\
MovRec                       & 1.013  & 1.015  & 1.032 & 1.013   & 1.012  & 0.999  & 1.004   & 0.994  & 0.990  & 1.113       & 1.742       & 4.207       \\
InfExp                       & 1.005  & 1.001  & 1.000 & 1.000   & 1.000  & 1.000  & 0.998   & 0.997  & 1.000  & 1.017       & 1.003       & 1.028       \\ \hline
\end{tabular}
\end{table*}

Recently, combinatorial optimization has seen a great deal of interest in the development of efficient heuristic algorithms \cite{bello2016neural, dai2017learning, barrett2020exploratory, cappart2020combining, yao2021reversible}. The deep Q-learning (DQN) is trained on a wide range of combinatorial problem instances and obtains a feasible solution for various applications. Even though \cite{manchanda2020gcomb, dai2017learning, bello2016neural, yao2021reversible, cappart2020combining, li2018combinatorial} provide high-quality solutions, only a select few can effectively perform the exploration behavior on the solution space, similar to local search \cite{barrett2020exploratory, yao2021reversible, li2018combinatorial}. These DQN-based models \cite{barrett2020exploratory, yao2021reversible, li2018combinatorial} primarily utilize graph neural networks (GNN), such as graph convolution network (GCN) \cite{li2018combinatorial}, message-passing neural network (MPNN) \cite{dai2017learning, barrett2020exploratory}, and graph attention network (GAN) \cite{cappart2020combining}. However, more recent works have proven that the GNNs inherently suffer from over-smoothing and information squashing \cite{li2018deeper, xu2018representation, zhou2020graph, chen2022bag}. These issues can be more intense in unsupervised tasks due to lacking supervision information \cite{yang2020toward}. In addition, the large message vectors restrict the scalability because of memory overhead. In light of the local search algorithms' performance and the limitation of GNNs, we study how is the performance of a lightweight model directly using node features in the cardinality-constrained maximization problems, then a natural question would be:
In light of the local search algorithms' performance and the limitation of GNNs, we study how is the performance of a lightweight model directly using node features in the cardinality-constrained maximization problems.
\textit{Is it possible to design a lightweight DQN model that can explore solution space like local search (LS) does and serve as a general-purpose algorithm for the combinatorial problem yet remain efficient in terms of runtime and memory consumption?}

\subsection*{\textbf{Contributions}}
\begin{itemize}
\item To combine the local search and heuristic exploration, we propose a general-purpose local search algorithm to heuristically explore the environment space to avoid additional traversal for better solutions. 

\item We introduce a DQN framework, \lssnn (\rels), to mimic the behavior of the aforementioned algorithm through reversible actions in a reinforcement learning (RL) environment. The agent uses a lightweight feed-forward network to heuristically explore the solution space and provide a feasible solution efficiently based on weighting the reduced number of state representation parameters in \cite{barrett2020exploratory}. Combined with a novel reward shaping that considers both objective value and the constraint can assist the agent to learn the exploration within constraint limitation. As a result, the model generalizes to a diverse set of applications without additional application-specific training and reduces memory usage.

\item Our approach is validated by conducting an empirical comparison between \lssnn, the MPNN-based local search model \eco \cite{barrett2020exploratory}, and the greedy local search algorithm (\grls) \cite{lee2009non}. The comparison is carried out across four diverse combinatorial problems. Our results show that \lssnn either performs equally well or better than \eco in terms of solution value and offers more efficient memory usage and runtime. Furthermore, our model offers an order-of-magnitude improvement in speed compared to \grls with a marginal difference in solution value.
\end{itemize}

\begin{figure*}[t]
    \centering
    \includegraphics[width=1.0\linewidth]{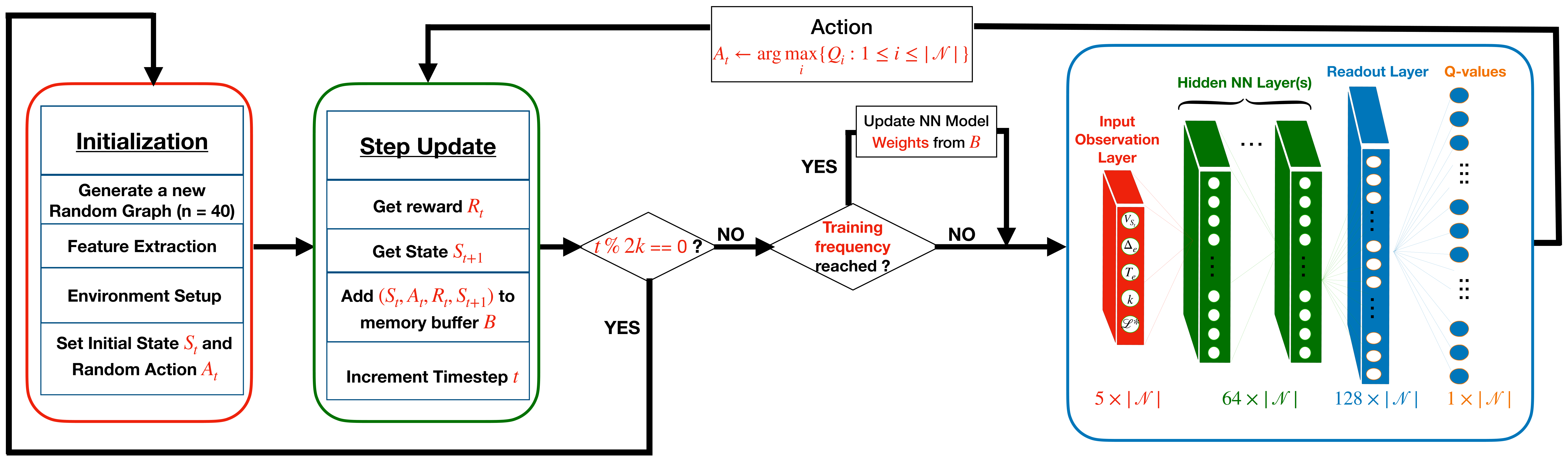}
    \hfill   
\caption{Overview of \lssnn training: 1) We generate a random ER graph with fixed size and initialize the environment for each training episode, where $t$ labels the iteration of timesteps;  2) Each element's parameters (denoted $x_e = \{V_{\{e\}}, \Delta_e,\mathcal T_e, V_{k_e}, \mathcal{L}^*\}$, Section~\ref{sec:Obs}) are updated and passed to the feed-forward network to calculate the Q-value, where the numbers below each layer are input/output vector dimension, and $|\mathcal{N}|$ is the ground set size; 3) Based on the $\epsilon$-greedy action rate of training, the agent's action $A_t$ will either be randomly chosen (exploration) or through a greedy-policy (exploitation, $A_t \gets \arg\max_i\{Q_i\}$); 4) Using action $A_t$, get the reward $R_t$ and the new state $S_{t+1}$ to update the environment and the network (if training frequency is reached); 5) Every $2k$ timesteps, we generate a new graph and repeat Steps (1) - (4) until the total timesteps $t=$ 1 million.}
  \label{fig0} 
  \vspace{-1.5em}
\end{figure*}

\section{Related work}
The advancements in AI techniques introduced new heuristic approaches based on deep learning \cite{bello2016neural, solozabal2020constrained, cappart2020combining, mazyavkina2021reinforcement, falkner2022learning, kool2022deep, kwon2021matrix}. One of the techniques is GNN, which has gained attention in solving graph-based combinatorial problems by aggregating information from the neighborhood to update node representations. This recursive aggregation scheme based on neighborhood relationships has proven to be a powerful technique in processing the representation of graphs \cite{dai2017learning, xu2018powerful, barrett2020exploratory, manchanda2020gcomb, cappart2020combining, falkner2022learning, kool2022deep, kwon2021matrix}. In general, these GNN-based approaches require either one-hot encoded vectors \cite{dai2017learning, kwon2021matrix, cappart2020combining} or node feature matrix \cite{barrett2020exploratory, manchanda2020gcomb, falkner2022learning, kool2022deep} as initial embeddings, which can be smoothed by the neighborhood aggregation to carrier graph structure information. However, recent studies have highlighted the inherent over-smoothing and information loss issues of GNNs \cite{li2018deeper, xu2018representation, zhou2020graph, chen2022bag}. This over-smoothing problem can also affect solving combinatorial optimization problems, as the goal is to identify distinguished nodes that maximize the objective function. 

Reinforcement Learning combined with GNNs is becoming increasingly popular for solving CO problems. Most notably, \cite{dai2017learning} proposed \svdqn, a GNN-based reinforcement learning approach, which uses the neighborhood aggregation to update the node representation and learns a greedy policy using Q-learning with the graph embedding vector. However, the state of \svdqn only considers the adjacency matrix and the state of nodes, and simply follows greedy policy at every step, which results in subpar performance \cite{li2018combinatorial}. As data size continuously increases, developing models that can scale to modern dataset sizes is imperative. Significant improvement in scalability can be seen in \gcomb \cite{manchanda2020gcomb}. \gcomb uses a probabilistic greedy mechanism to predict the quality of the nodes by a trained GCN. By evaluating the quality of nodes, \gcomb is able to prune those unlikely to be in the solution set, allowing it to generalize to large graphs of millions or billions of nodes while maintaining the performance of \svdqn on Maximum Coverage on the bipartite graph (MCP), Maximum Vertex Cover (MVC), and Influence Maximization (IM). However, similar to \svdqn, it does not involve local search behavior resulting in its performance being marginally lower than \svdqn.

Therefore, \cite{barrett2020exploratory} proposes \eco, an improvement over \svdqn by allowing the agent to explore, i.e., the agent allows performing both adding or removing actions, known as reversible action. The removal action allows the agent to step back, re-evaluate the states, and improve solutions better than the greedy. Their demonstration shows that the \eco outperforms \svdqn and MaxCutApprox algorithm \cite{barrett2020exploratory} (greedy-based) in the MaxCut problem with negative edge weight. The model uses the normalized observations of nodes as messages and the neighborhood normalization for aggregation, which are beneficial for node-specific tasks and for cases where the observation values vary significantly. However, the normalized node features not only lead to information loss after several iterations of message passing but also limited the model to application-specific problems. Although \eco has demonstrated good performance, it encounters memory scaling challenges when applied to large and dense graphs, particularly when the message has high-dimensional vector space computed by an initial embedding layer. 

In addition, more recent works have investigated local search techniques to address the combinatorial optimization problem. For example, \cite{li2018combinatorial} proposed an approach based on tree search, which utilizes GCN to present a probability map and then refine the solution with a local search algorithm. This approach requires iterations over all vertices to find a replacement, which is an inevitable step in the local search. In a study by \cite{falkner2022learning}, they formulate a scheduling problem with sequential selection as a Markov Decision Process (MDP) and use RL to learn a parameterized policy with local search actions employing the GNN-based encoder. The trained RL can learn the local search by acceptance, neighborhood selection, and perturbations, yielding additional iterations like local search algorithms. Another work related to our approach is \cite{yao2021reversible}, which uses reinforcement learning with MPNN to adopt an expensive swapping search in size of $O(n^2)$. Overall, these works show that even though the results of \cite{dai2017learning, barrett2020exploratory, li2018combinatorial, yao2021reversible, falkner2022learning} show a promising prospect of finding better solutions than greedy, the local search and MPNN methods remain challenges in solving combinatorial optimization problems. The proposed model \lssnn explores the possibility of the RL agent learning a heuristic exploration strategy without relying on message passing and if it can generalize to various cardinality-constrained combinatorial optimization problems.

\section{Preliminaries} \label{sec:prelim}
\textbf{RL Encoding:}
The combinatorial optimization problem is defined in the standard RL framework as follows:
\begin{itemize}
    \item States $S$: is a vector of the environment representation which usually includes graph embeddings \cite{dai2017learning, barrett2020exploratory}. In our model, the state is a sequence of elements on graph $G$, and its embedding representation is a set of element-specific parameters $\{x_{e_1}, x_{e_2},\dots, x_{e_n}\}$ (Section~\ref{sec:Obs}). It is easy to see that this representation of the state can be used across different applications as each parameter is weighted the same by the pre-trained network model.

    \item Actions $a$: is an element $e$ from $\mathcal{N}$, which adds or removes this element from the solution set by flipping the incumbent state $V_{\{e\}}$ (Section~\ref{sec:Obs}, the value is flipped to 1 if $e \notin V$ and to 0 if $e \in V$, where $V$ is the partial solution of current state $S$), and updates other parameters in $x_e$.
    
    \item Rewards $R(S, a)$: is defined as the difference in the objective function after updating the state \cite{dai2017learning}, which is $R(S, a)=f(S') - f(S)$, where $S'$ is the state after current state $S$ taking action $a$. In this case, the cumulative reward $R$ at the terminal state is the objective function value. In our model, the reward is shaped to incorporate constraint in Section~\ref{sec:Reward}.
    
    \item Transition $(S, a, R, S')$: uses tuple to present the process of moving from $S$ to $S'$ with giving action $a$.
\end{itemize}

\textbf{Double Q-learning:}
Double Deep Q-network (Double DQN) \cite{van2016deep} is an off-policy reinforcement learning algorithm that reduces the overestimation of DQN to improve the accuracy of value estimation. A target network with the same architecture as the original DQN is periodically updated to obtain the Q-value. The Q-value function of the target network is defined as follows:
\small
\begin{equation}
    Q^{target} \equiv R(S, a) + \gamma Q(S', \underset{a}{\arg\max} \ Q(S',a;\theta),\theta^{target}),
\end{equation}
\normalsize
where $\theta$ is the online weights updated by $S$, $\theta^{target}$ is the target network weights updated by $S'$, and $\gamma\in[0,1)$ is the discount factor.

\begin{algorithm}[t]
   \caption{ReversibleLocalSearch$ (f, \mathcal N, k, \lambda_{\Delta}, \lambda_t)$} \label{algo:RLS}
  \begin{algorithmic}[1]
     \State {\bfseries Input:} evaluation oracle $ f:2^{\mathcal N} \to \mathbb{R}$, constraint $k$, marginal gain influence weight $\lambda_{\Delta}$, element age influence weight $\lambda_t$
     \State Let $V \gets \emptyset $, $V^* \gets \emptyset$, $t=0$
     \State $f(e|V) =  f(V \cup \{e\}) - f(V)$  if $e \notin V$,  otherwise $f(V$ \textbackslash $ \{e\}) - f(V)$
     \State $\mathcal T \gets \{t_e = 0 $ ; $ e \in \mathcal N\}$
     \label{line:TInit}
     \For{$t = 1$ to $2k$ } \label{line:timestep}
       \State $\mathcal T \gets \{t_e + 1 $ ; $ e \in \mathcal N\}$
       \State $\Delta \gets \{f(e|V)$ : $e \in \mathcal N \}$
       \State $x \gets \arg\max_e\{\lambda_{\Delta}\cdot\Delta_e +  \lambda_t\cdot  \mathcal T_e$ ; $ e \in \mathcal N \}$
       \If{$\Delta_x > 0$ or $|V| \leq k$} \label{line:exp2}
            \If{$x \notin V$}
                \State $V \gets V \cup \{x\}$
            \Else
                \State $V \gets V$ \textbackslash $\{x\}$
            \EndIf
            \State $\mathcal T_x \gets 0$
            \State $V^* \gets \arg\max \{ f(V^*), f(V) \} $ \label{line:exp3}
        \Else
            \State $x \gets \arg\max_e\{\lambda_{\Delta}\cdot\Delta_e +  \lambda_t\cdot  \mathcal T_x$ ; $e \in V \}$
            \label{line:exp0}
            \State $V \gets V$ \textbackslash $\{x\}$ \label{line:exp1}
        \EndIf
     \EndFor
     
     \State \textbf{return} \textit{$V^*$}
  \end{algorithmic}
\end{algorithm}

\section{RELS-DQN}
\label{lssnn}
To start, we modified the local search algorithm \cite{bian2017guarantees} to allow it to explore unvisited nodes after a few iterations instead of extra queries after each selection of the maximum singleton. In Algorithm \ref{algo:RLS}, we provide two parameters to balance the weight of exploration and greedy selection. For each node, we set up an element-wise initialized timer $\mathcal{T}$ (line \ref{line:TInit}) to track if the node has been visited yet. The $2k$ iterations imply two stages of selection. In the first $k$ iterations, the marginal gain of the objective function $\delta_e$ is selected greedily as the $\mathcal{T}_e$ remains the same for $e \notin V$. As for the second $k$ iteration, the unvisited nodes have a larger weight on $\mathcal{T}_e$ to encourage exploring new nodes instead of following the margin gain. However, it is infeasible to find the best weight parameters that apply to various applications. 

Next, we propose \lssnn inspired by \cite{barrett2020exploratory}. Our model mimics the behavior of the general-purpose local search algorithm defined in Algorithm \ref{algo:RLS}, and balances the greedy policy and exploration through a network model instead of weight parameters. \lssnn modified the environment observations in \cite{barrett2020exploratory}. We provide element-wise defined parameters (Section \ref{sec:Obs}) as the representation of the environment state. Our model uses a lightweight feed-forward network (Section \ref{sec:network}) that provides the flexibility to learn an application. This network can apply to different applications while remaining efficient in terms of memory and runtime. As for the reward shaping (Section \ref{sec:Reward}), we use the cardinality constraint $k$ as part of reward shaping to enable exploration of the local optimal solution within the constraint limitation. Figure \ref{fig0} illustrates the training framework of \lssnn, where it explores small randomly generated synthetic graphs to learn how to obtain solutions through local search behavior.

\subsection{Element Parameters} \label{sec:Obs}
In this section, a set of parameters are defined at the element level to describe the current state of the environment. These parameters provide the feature representation to the RL model for balancing exploration, exploiting the greedy policy, and taking into account constraint limitation. Existing models such as \svdqn \cite{dai2017learning} and \eco \cite{barrett2020exploratory} use one and seven element-specific parameters respectively for solution selection. While these are performant models, their lack of constraint parameters hinders their ability to handle constrained problems. In addition, \svdqn only uses one element parameter which restricts its local search capability, while the seven parameters of \eco lead to redundancy due to its global observations. Therefore, we list the parameters used in our model as follows:
\begin{itemize}
    \item \textbf{Incumbent State $V_{\{e\}}$} is a binary bit to indicate if the element is selected or not in current state $S$, such that it is 1 if $e \in V$; otherwise 0. This parameter of the entire set is denoted as $V_{S}$, such that the partial solution $V$ is denoted as $V=\{x\in V_S | x=1\}$.
    \item \textbf{Marginal Gain $\Delta_e$} is the gain/loss of adding/removing element $e \in \mathcal N$ from the partial solution $V$. This parameter encourages the model to maximize the gain regardless of whether the action is removing or adding an element.
    \item \textbf{Inactive Age $\mathcal T_e$} provides the timestep that element $e \in \mathcal N$ has not been selected as the action $a$. It is reset to 0 every time $a \gets \{e\}$, which discourages the model from selecting the same action every iteration and aid in extensive exploration of the solution space.
    \item \textbf{Feasible candidate $V_{k_e}$} provides the permissible actions of an element. For the entire set, its candidate state is represented as $V_{k}$. This parameter discourages the model to generate a solution set size larger than $k$ and encourages removing action when it exceeds constraint limitation. For this parameter, all elements are set to 1 if $|V|\leq k$; otherwise, only the elements in the partial solution set are 1, and the rest elements are 0.
    \item \textbf{Distance to local optimum $\mathcal L^*$} is the distance of current $f(S)$ to the best observed $f(S^*)$. The $S^*$ is the state of the best solution that is updated when $f(S) > f(S^*)$ at the end of each timestep. As the actions reducing solution value are possible, we need this parameter to track the best solution that has been encountered during the episode. 
\end{itemize}

\subsection{Network Architecture}
\label{sec:network}
The DQN network architecture determines how the Q-value for the action policy is calculated. Existing literature has primarily focused on GCN \cite{kipf2016semi, vesselinova2020learning, manchanda2020gcomb} and MPNN \cite{dai2017learning, barrett2020exploratory} to determine the best action based on the updated node embedding from the neighborhood aggregation. While the GNN aggregates information from local graph neighborhoods and the embeddings are encoded with the multi-hop neighbors, it also smooths the features that are well-designed to intuitively reflect the environment state. Especially, the parameters defined as the environment state in section~\ref{sec:Obs} are additional information that is not attached to nodes and edges and should avoid any operation that can cause information loss, such as neighborhood normalization \cite{hamilton2020graph}. Therefore, these advantages of GNN adversely impact the robustness and ability to solve various combinatorial optimization problems. Additionally, the memory overhead of MPNN and GCN network architecture can limit their scalability to large data instances. \lssnn addresses the robustness and scalability concerns of the MPNN-based models by utilizing two fully connected feed-forward layers with \relu activation functions represented as follows:
\begin{equation}
\label{lsnn}
\mu _e= \relu(\theta_2~\relu(\theta_1 x_e)),
\end{equation}
where $x_e = \{V_{\{e\}}, \Delta_e,\mathcal T_e, V_{k_e}, \mathcal{L}^*\} \in \mathbb{R}^5$ is the feature vector of a single element $e\in \mathcal{N}$, and $\mu$ is its output vector from the hidden layers. $\theta_1 \in \mathbb{R}^{m\times 5}, \theta_2 \in \mathbb{R}^{m\times m}$ are the weight of the corresponding layers, and $m$ is the number of neurons in the hidden layers. 

The feed-forward layers directly capture information about the environment representations and use its weight matrix and \relu activation function to produce the hidden features. Then, the readout layer produces the Q-value from each element's hidden feature $\mu_e$ and the pooled embedding over the ground set as follows:
\begin{equation}
Q_e = \theta_4[\relu(\theta_3\frac{1}{|\mathcal{N}|}\sum _{u\in \mathcal{N}}\mu_u),\mu_e],
\end{equation}
where $\theta_3 \in \mathbb{R}^{m \times m}$, $\theta_4 \in \mathbb{R}^{2m}$, $\left[ \cdot,\cdot \right]$ is the concatenation operator, and $|\mathcal{N}|$ represents the number of elements in ground set. The embedding of the ground set $\mu_u, u\in \mathcal{N}$ uses the shared $\mu_e$, which is later normalized with $|\mathcal{N}|$ and fed into a linear pooling layer with \relu activation function. The concatenation of the $\mu_e$ and the pooled embedding over the ground set, $\relu(\theta_3\frac{1}{|\mathcal{N}|}\sum _{u\in \mathcal{N}}\mu_u)$, is used to extract the Q-value so that the model selects the action $a \gets \arg\max_i\{Q_i: 1\le i \le |\mathcal N|\}$.

\subsection{Reward reshaping} \label{sec:Reward}
The reward reshaping of \lssnn primarily consists of the state reward $F(V_{S})$ and the constraint reward $D(V_{S})$ in order to find the best solution within constraints by exploration. Given the state reward $F(V_{S}) = max(f(S) - f(S^*), 0)$ and the constraint reward $D(V_{S}) = k - |V|$, the non-negative reward for solving constrained instances is formally given as follows:
\small
\begin{equation}
R(S, a )=\begin{cases}
\hfil max\{ \frac{F(V_{S})  \times D(V_{S})}{|\mathcal{N}|},0\} & a \notin V, |V| \ge k
\\
\hfil max\{\frac{F(V_{S})}{|\mathcal{N}|},0\} & \text{, otherwise}
\end{cases}
\end{equation} 
\normalsize
where $V_{S}$ is the incumbent state at state $S$; $S^*$ is the best observed state up to current state and $k$ is the cardinality constraint. 

The state reward $F$ provides a positive reward when an action improves the objective value over the best-observed state $S^*$ and does not receive a penalty when the action reduces the solution value. On the other hand, the constraint reward $D$ works as follows: if the size of the solution is within the constraint limitation $|V| < k$, the agent receives the state reward; if $|V| \ge k$, the agent is encouraged to remove an element $e$ from the partial solution $V$ by adding a penalty to the adding action. Additionally, the exploration is considered by removing all penalties and not rewarding the agent unless the solution is improved. Therefore, reward shaping considers two scenarios: if the agent takes any actions within the constraint limit or the removing actions at $|V|\ge k$, the environment provides the state reward; if the agent takes the adding action at $|V|\ge k$, the environment changes any positive reward into zero by providing the constraint penalty $D<0$. In this case, the environment only provides a positive reward when a better solution is found within the constraint, and the non-negative reward allows the agent to explore actions that can reduce the solution value. 

\section{Empirical Evaluation}
In this section, we analyze the effectiveness and efficiency of \lssnn as a general purpose algorithm. We provide the source code \href{https://drive.google.com/file/d/1cQDaj1ZdlQMYZC-NfBdnBwqspSzbjVkt/view}{link}\footnote{Code Repository: \url{https://drive.google.com/file/d/1cQDaj1ZdlQMYZC-NfBdnBwqspSzbjVkt/view}} that includes experimental scripts and datasets. With our empirical evaluations, we address the following questions:

\textbf{Question 1:} Is a shallow network architecture better at solving CO problems than deep neural networks?

\textbf{Question 2:} Can the \lssnn model solve a diverse set of problems without application-specific training?

\textbf{Question 3:} How efficient is \lssnn compared to the MPNN-based \eco and \grls? 
\label{expset}

\subsection{Experiment setup}
\label{exp-setup}
\textbf{Applications:} Our experiment set consists of a set of four diverse combinatorial optimization problems: Maximum Cut (MaxCut) \cite{barrett2020exploratory}, Directed Vertex Cover with Costs (MaxCov) \cite{harshaw2019submodular}, Movie Recommendation (MovRec) \cite{fahrbach2019non}, and Influence-and-Exploit Marketing (InfExp) \cite{amanatidis2020fast}. Details of all applications and their objective functions are defined as follows:

\textbf{Maximum Cut Problem} (MaxCut) is defined as follows. For a given graph $G(\mathcal{N}, E, w)$, where $\mathcal{N}$ represents the set of vertices, $E$ denotes the set of edges, and $w:E\rightarrow \mathbb{R}^*$ represents the weight of edges with $w \in \{0, \pm 1\}$. Find a subset of nodes $V \subseteq  \mathcal{N}$ such that the objective function is maximized: \cite{barrett2020exploratory}
\begin{equation}
f(V):=max \sum_{u \in V, v \in \overline{V}} w(u, v),
\end{equation}
where $\overline{V}=\mathcal{N} \setminus V$ represents the complementary set. In addition, the constrained MaxCut problem projects to $|V|\leq k$, in which $0 < k \leq |\mathcal{N}|$ is a constraint. 

\textbf{Directed Vertex Cover with Costs} (MaxCov) is defined as follows. For a given directed graph $G(\mathcal{N}, E, w)$, let $w:E\rightarrow \mathbb{R}$. For a subset of nodes $V \subseteq  \mathcal{N}$, the weighted directed vertex cover function is $g(V)=\sum_{u \in N(V) \cup V} w_u$, where $N(V)$ represents the set of nodes pointed by the subset $V$. \cite{harshaw2019submodular} assumes that there is a nonnegative cost $c_n$ for each $n \in \mathcal{N}$. Therefore, the objective function is defined as \cite{harshaw2019submodular}:
\begin{equation}
g(V)-c(V):=\sum_{u \in N(V) \cup V} w_u - \sum_{u \in V} c_u,
\end{equation}
where the cost associated with each node is set to $c(n) = 1+max\{d(n)-q,0\}$ for a fixed cost factor $q=6$ and the out-degree $d(n)$ of $n \in \mathcal{N}$.

\textbf{Movie Recommendation} (MovRec) produces a subset from a list of movies for a user based on other users with similar ratings. We use the objective function in \cite{fahrbach2019non} defined as follows: 
\begin{equation}
f(V):= \sum_{i \in \mathcal{N}} \sum_{ j \in V} s(i, j) -\lambda \sum_{i \in V} \sum_{ j \in V} s(i, j)
\end{equation}
Where $s_{ij}$ is the inner product of the ratings vector between movie $i$ and $j$, and $\lambda=5$ is the diversity coefficient which the large $\lambda$ penalizes similarity more \cite{amanatidis2020fast}. 

\textbf{Influence-and-Exploit Marketing} (InfExp) consider how the buyers of a social network can influence others to buy goods for revenue maximization. For a given set of buyer $\mathcal{N}$, each buyer $i \in \mathcal{N}$ is associated with a non-negative concave function $f_i(x)=a_i\sqrt{x}$ where the $a_i$ a number followed by Pareto Type II distribution with $\lambda=1, \alpha=2$ \cite{amanatidis2020fast}. For a given subset $V \subseteq \mathcal{N} \setminus  \{i\}$ of buyer $i$, the objective function is defined as follows \cite{amanatidis2020fast}:
\begin{equation}
f(V) := \sum_{i \in \mathcal{N} \setminus V} f_i(\sum_{j \in V \cup \{i\}} w_{ij})
\end{equation}
where the weight of the graph is uniformly randomly generated from $U(0,1)$.

\begin{table}[h]
    \centering
    \caption{Dataset and graph size.}\label{tab:dataset}
    \begin{tabular}{l p{32mm} l l}
      \hline\hline 
      Application & Dataset & $|\mathcal{N}|$ & $|E|$ \\ [0.5ex] 
      \hline 
      MaxCut & ego-Facebook(fb4k) & 4039 & 88234 \\
      MaxCut & musae-FB(fb22k) & 22470 & 171002 \\
      MaxCov & email-Eu-core(eu1k) & 1005 & 25571 \\
      MovRec & MovieLens(mov1.7k) & 1682 & 983206 \\
      InfExp & BarabásiAlbert(ba300) & 300 & 1184 \\
      \hline 
    \end{tabular} \\
\end{table}

\textbf{Benchmarks:} We compare \lssnn to the MPNN heuristic framework \eco \cite{barrett2020exploratory}, the standard greedy algorithm (\sg), the vanilla local search algorithm (\grev) and the local search algorithm\cite{lee2009non} (\grls). The \eco serves as a benchmark after adapting the feasible candidate $V_k$ in environment observation and the constrained reward function. \grev is a greedy algorithm that allows reversible actions to arrive at a better value until no further improvement is possible. \grls meanwhile repeats itself within the cardinality constraint $k$, checking if the existing nodes satisfy the credentials for removing or swapping after every new addition. Additionally, we use sparse representation on MPNN using Torch-Geometric to evaluate memory improvement.

\textbf{Training:} \lssnn model is trained on MaxCut only, and \eco models are trained on specific applications. \eco model is modified and adapted to the constrained problem by applying feasible candidate parameter $V_k$ (Section \ref{sec:Obs}) and reward shaping(Section \ref{sec:Reward}). We train all models using synthetic graphs ($n$=40, $k$=30) with each episode $2k$ timesteps to make sure each graph is explored entirely and the element parameters are learned sufficiently. The synthetic training data is generated in the following ways:

\begin{itemize}
\item \textbf{MaxCut:} Erd\"{o}s-R\'{e}nyi graphs (ER) \cite{erdos1960evolution} with probability $p=0.15$ and edge weight $w_{ij}\in \{0, \pm 1\}$.
\item \textbf{MaxCov:} ER graphs with $p=0.15$ and $w_{ij}\in \{0, +1\}$. 
\item \textbf{MovRec:} ER graphs with $p=0.15$ and $w_{ij}\in (0,1)$ generated uniformly randomly.
\item \textbf{InfExp:} Barabasi-Albert graph (BA) \cite{albert2002statistical} with $m=4$ and $w_{ij}\in (0,1)$ generated uniformly randomly.
\end{itemize}
 
\textbf{Testing:} \lssnn model uses the pre-trained model of MaxCut to validate all applications and datasets. Conversely, the \eco model uses the application-specific pre-trained model for each problem. We evaluate the performance using synthetic ER graph ($n$=200, $p$=0.03), BA graph ($n$=200, $m$=4) and large datasets listed in Table \ref{tab:dataset}. In addition, all applications are validated on a hundred graphs except the InfExp is validated on ten graphs.

\textbf{Hardware environment:}
We run all experiments on a server with a TITAN RTX GPU (24GB memory) and 80 CPU Intel(R) Xeon(R) Gold 5218R cores. 

\subsection{Architecture Evaluation}
\label{arch-eva}
\begin{figure}[h]
    \centering
    \caption{The learning curves of the RL models for MaxCut on ER-40 graph. \lssnn and \eco converge to a higher solution value than \svdqn while \lssnn reaches convergence earlier than \eco.}
    \subfloat[\label{fig:5a}]{
    \includegraphics[width=0.48\linewidth]{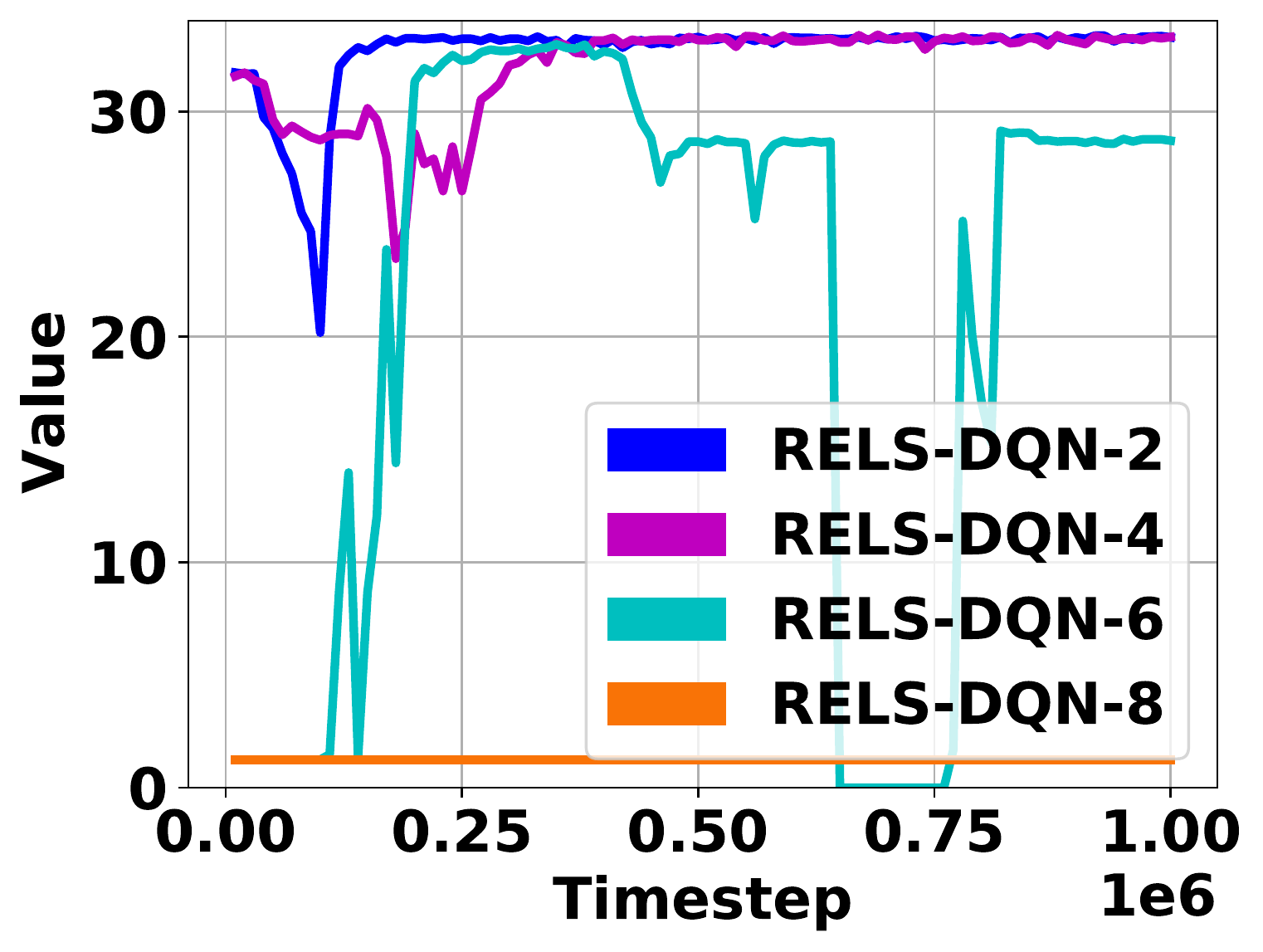}}
    \hfill
    \subfloat[\label{fig:5b}]{
    \includegraphics[width=0.48\linewidth]{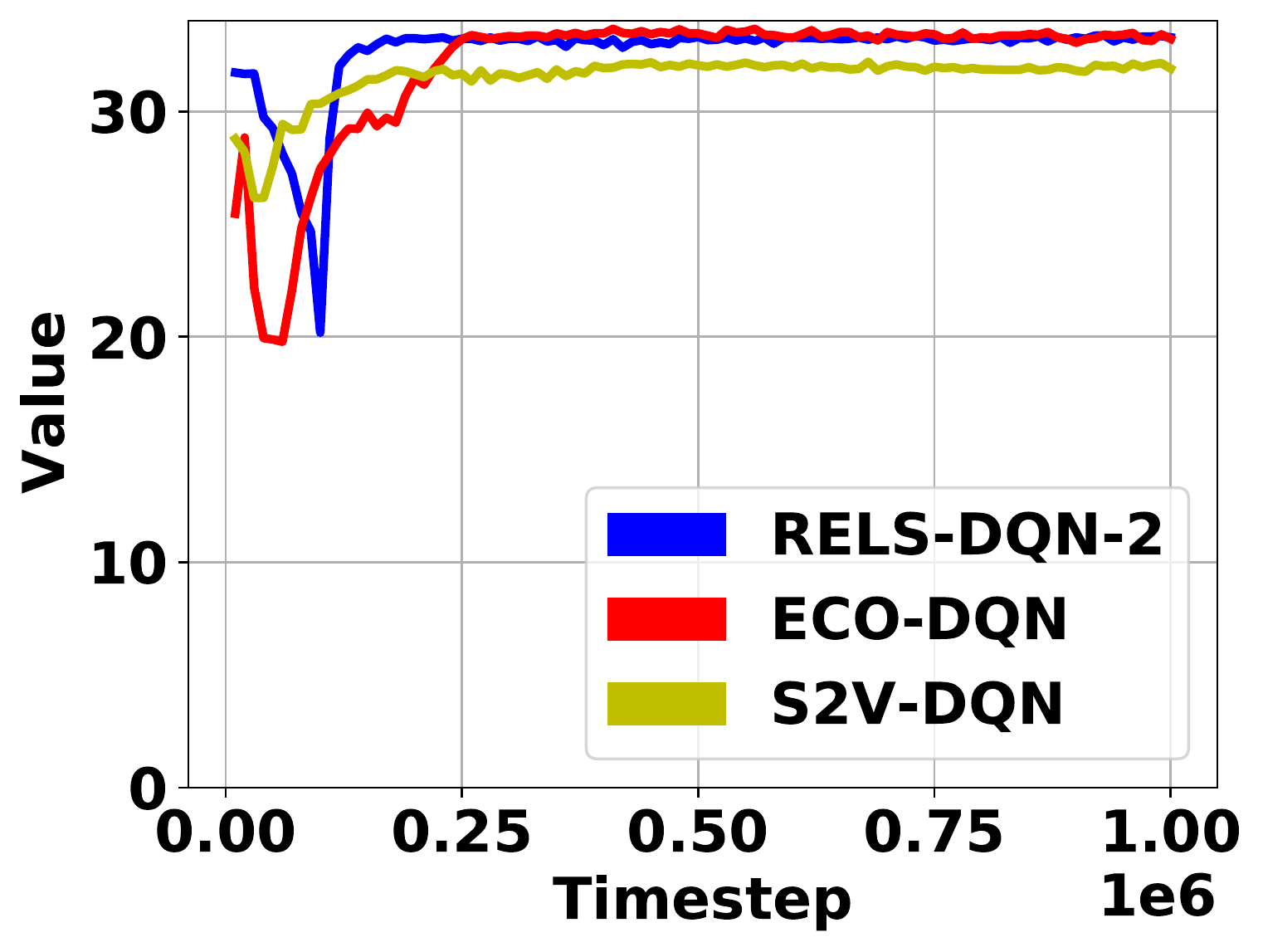}}
    \hfill
    \subfloat[\label{fig:5c}]{
    \includegraphics[width=0.48\linewidth]{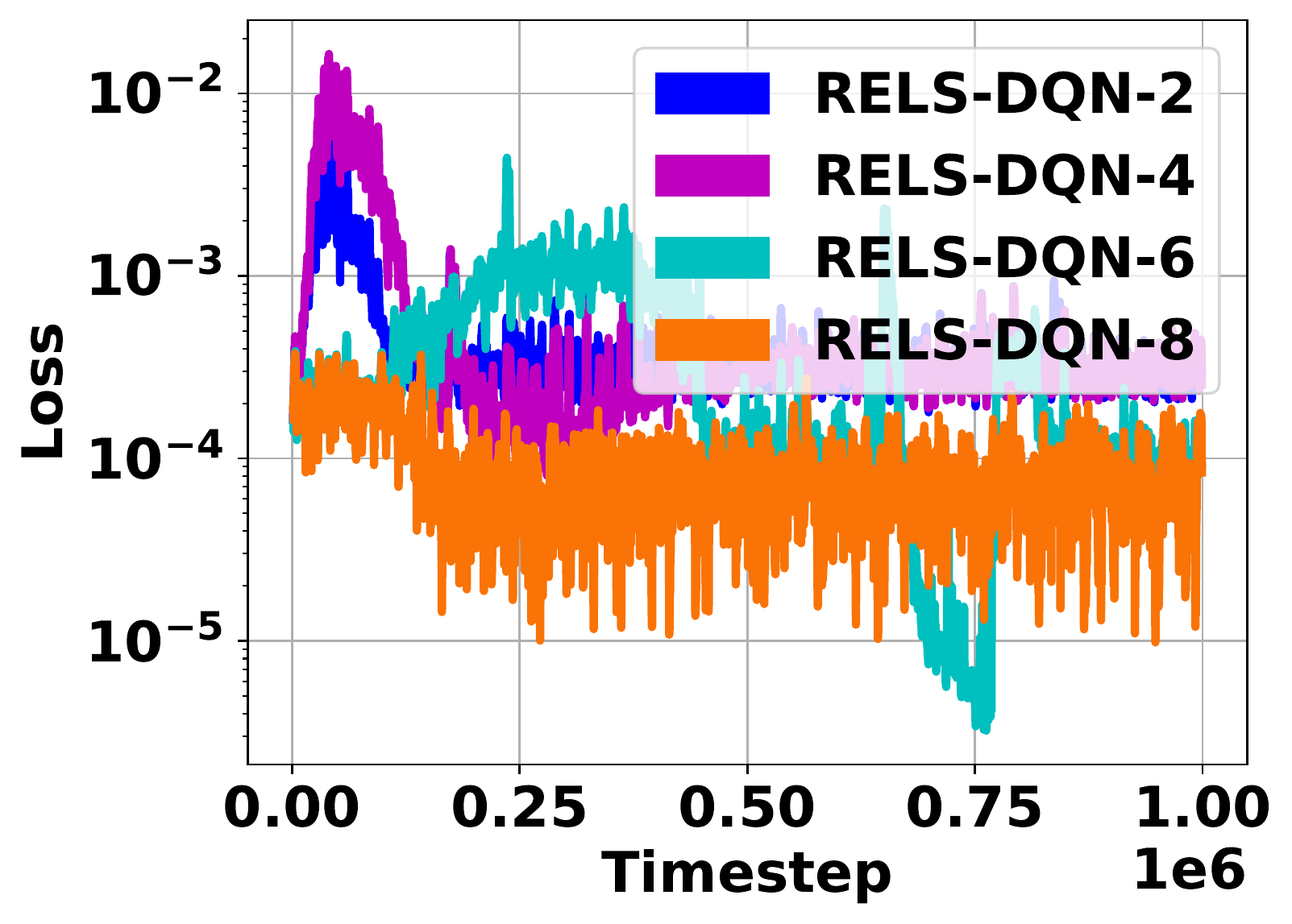}}
    \hfill
    \subfloat[\label{fig:5d}]{
    \includegraphics[width=0.48\linewidth]{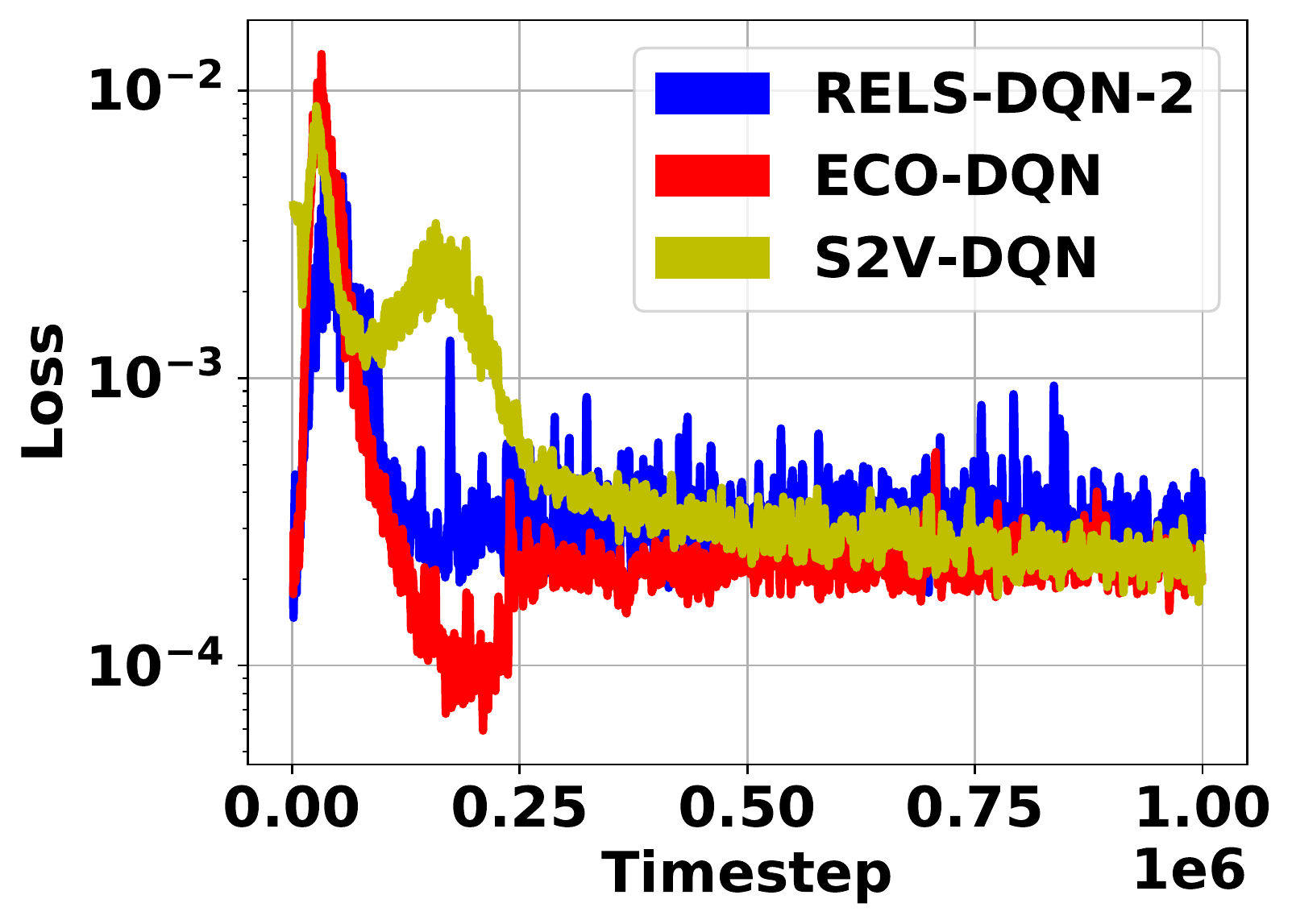}}
  \label{fig5} 
   \vspace{-1.5em}
\end{figure}

In this section, we describe the learning characteristics of the DQN models on the MaxCut problem. Figure \ref{fig:5a} compares the training performance of \lssnn by increasing the number of hidden layers in Equation \ref{lsnn} up to 8 layers. As shown, \lssnn-2 and \lssnn-4 converge to the highest solution value, and \lssnn-2 exhibits the ability to converge earlier. \lssnn-6 cannot converge to a high value, while \lssnn-8 shows the inability to learn the MaxCut problem. These results demonstrate the adverse effect of using a complex neural network architecture to solve combinatorial problems. Based on our evaluation, we choose \lssnn-2 as our model for all evaluations as it demonstrates the best learning ability. In figure \ref{fig:5b}, we observe that both \lssnn-2 and \eco converge to the best solution with ours converging earlier. \svdqn cannot achieve higher solution values due to its lack of local search. 

\begin{figure*}[t]
    \centering
    \caption{Ilustrates the testing results of \lssnn, \eco, \sg, \grev and \grls on four applications. \lssnn uses the pre-trained model of MaxCut (ER, $n=$40) for all applications, while \eco uses application-specific model for each application. The $timeout$ for each application is set to \textbf{24 hours}.}
  \subfloat[MaxCut (ER200) \label{fig_uncon_maxcutu}]{%
       \includegraphics[width=0.24\linewidth]{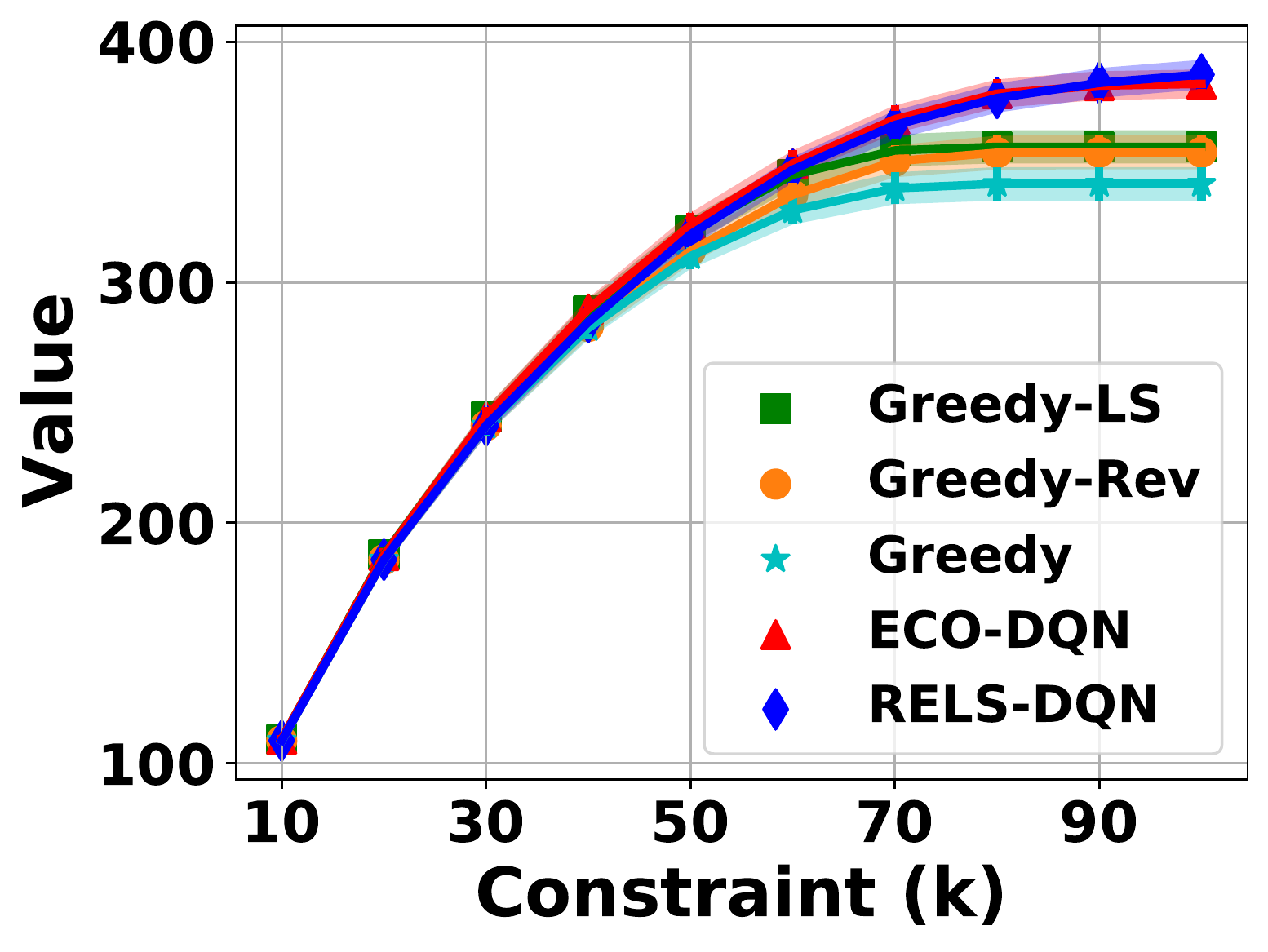}}
    \hfill
  \subfloat[MaxCov (ER200) \label{fig_uncon_maxcovu}]{%
        \includegraphics[width=0.24\linewidth]{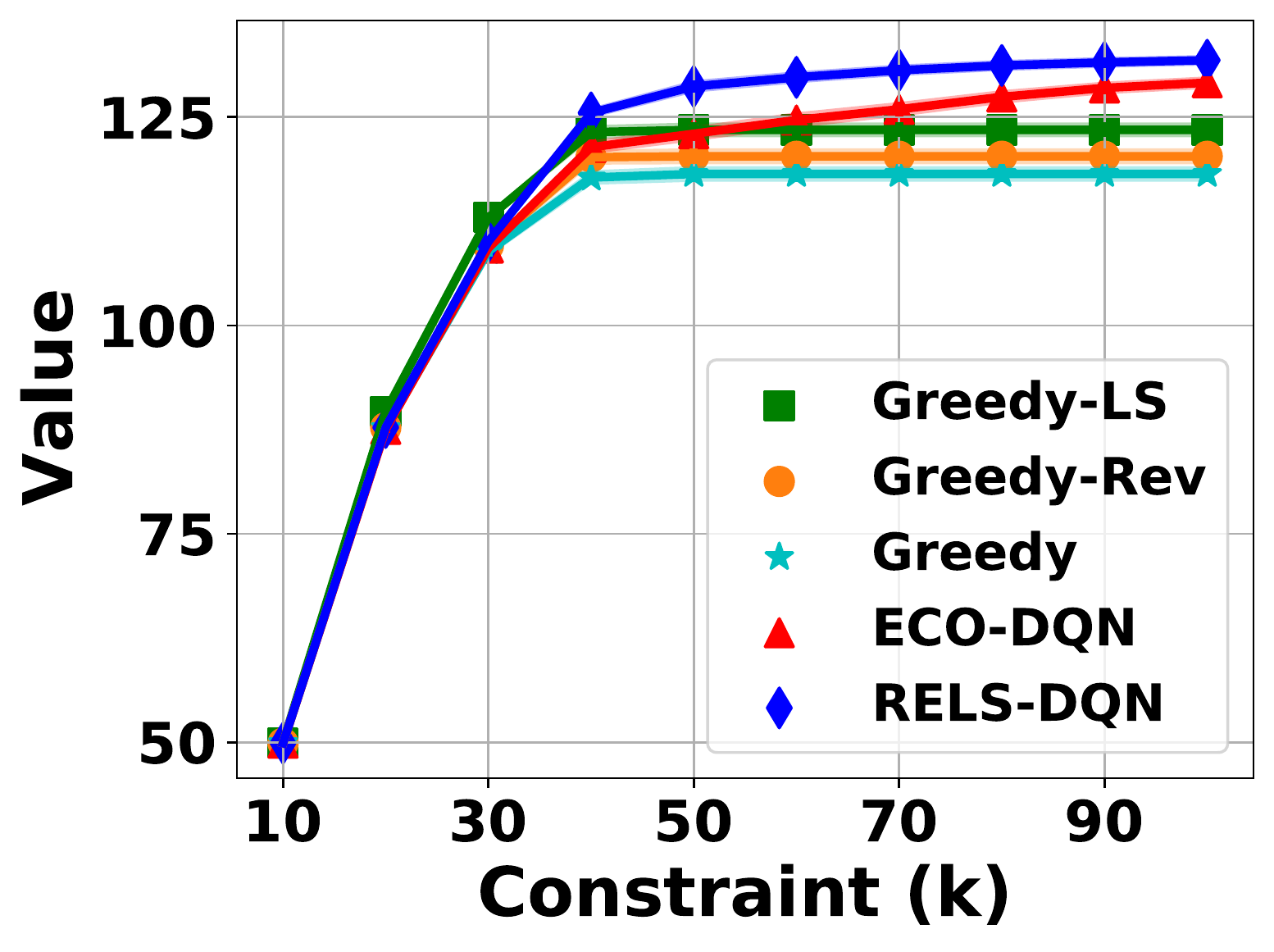}}
    \hfill   
  \subfloat[MovRec (ER200) \label{fig_uncon_movmaxu}]{%
        \includegraphics[width=0.24\linewidth]{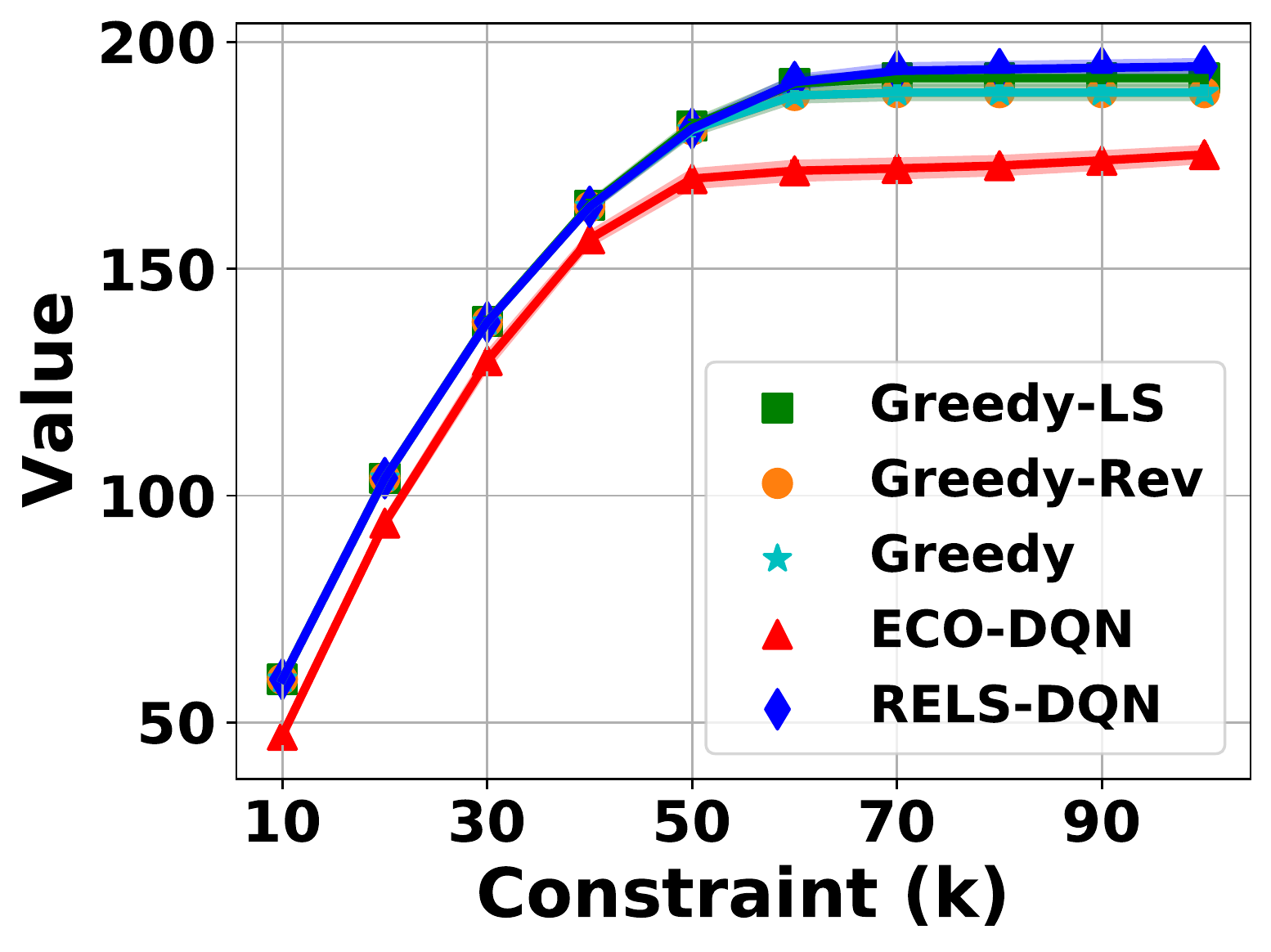}}
    \hfill   
  \subfloat[InfExp (BA200) \label{fig_uncon_infexpu}]{%
        \includegraphics[width=0.24\linewidth]{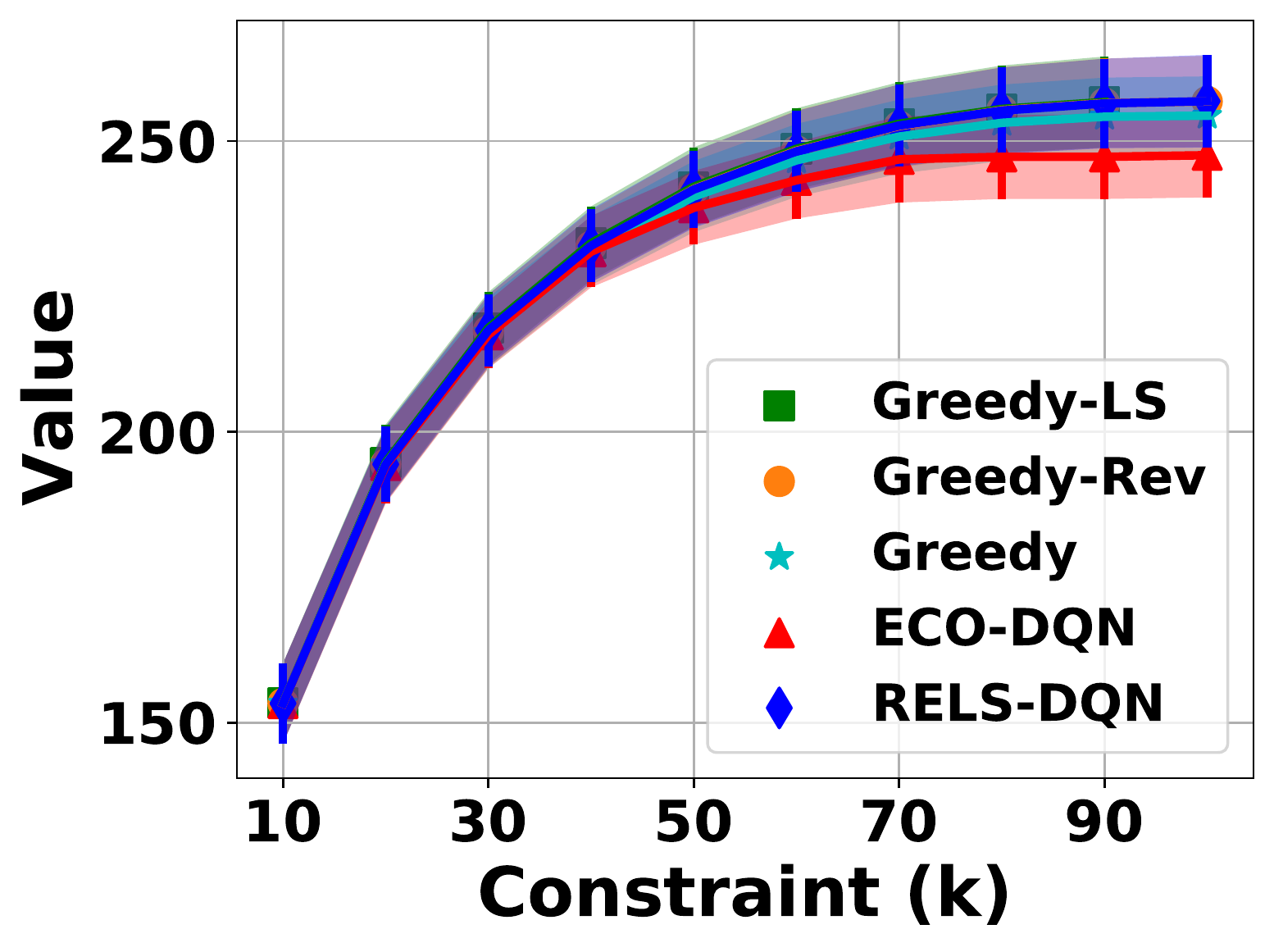}}
    \\
    \subfloat[MaxCut (BA200) \label{fig_uncon_maxcut2}]{%
       \includegraphics[width=0.24\linewidth]{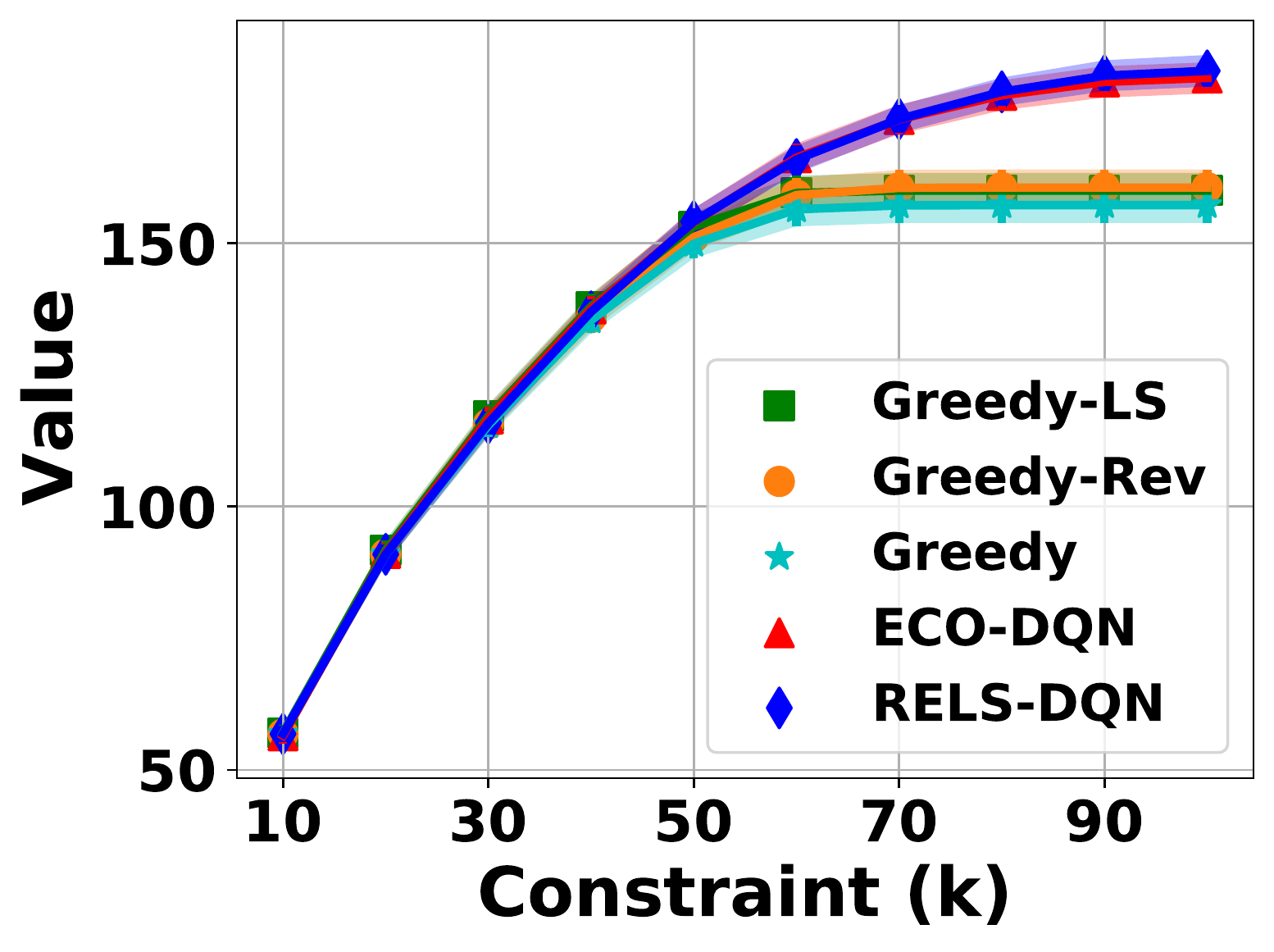}}
    \hfill
  \subfloat[MaxCov (BA200) \label{fig_uncon_maxcov2}]{%
        \includegraphics[width=0.24\linewidth]{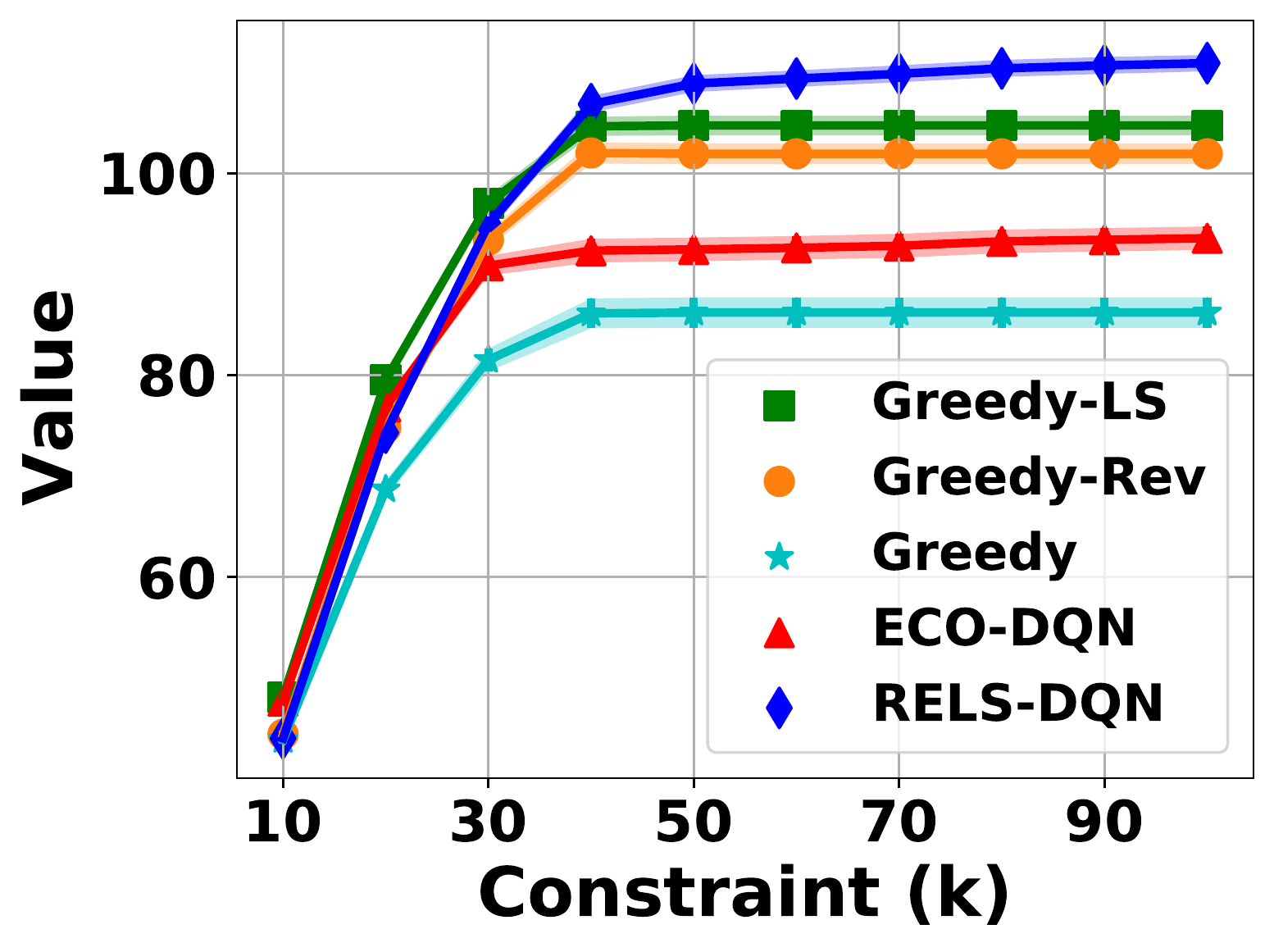}}
    \hfill   
  \subfloat[MovRec (BA200) \label{fig_uncon_movmax2}]{%
        \includegraphics[width=0.24\linewidth]{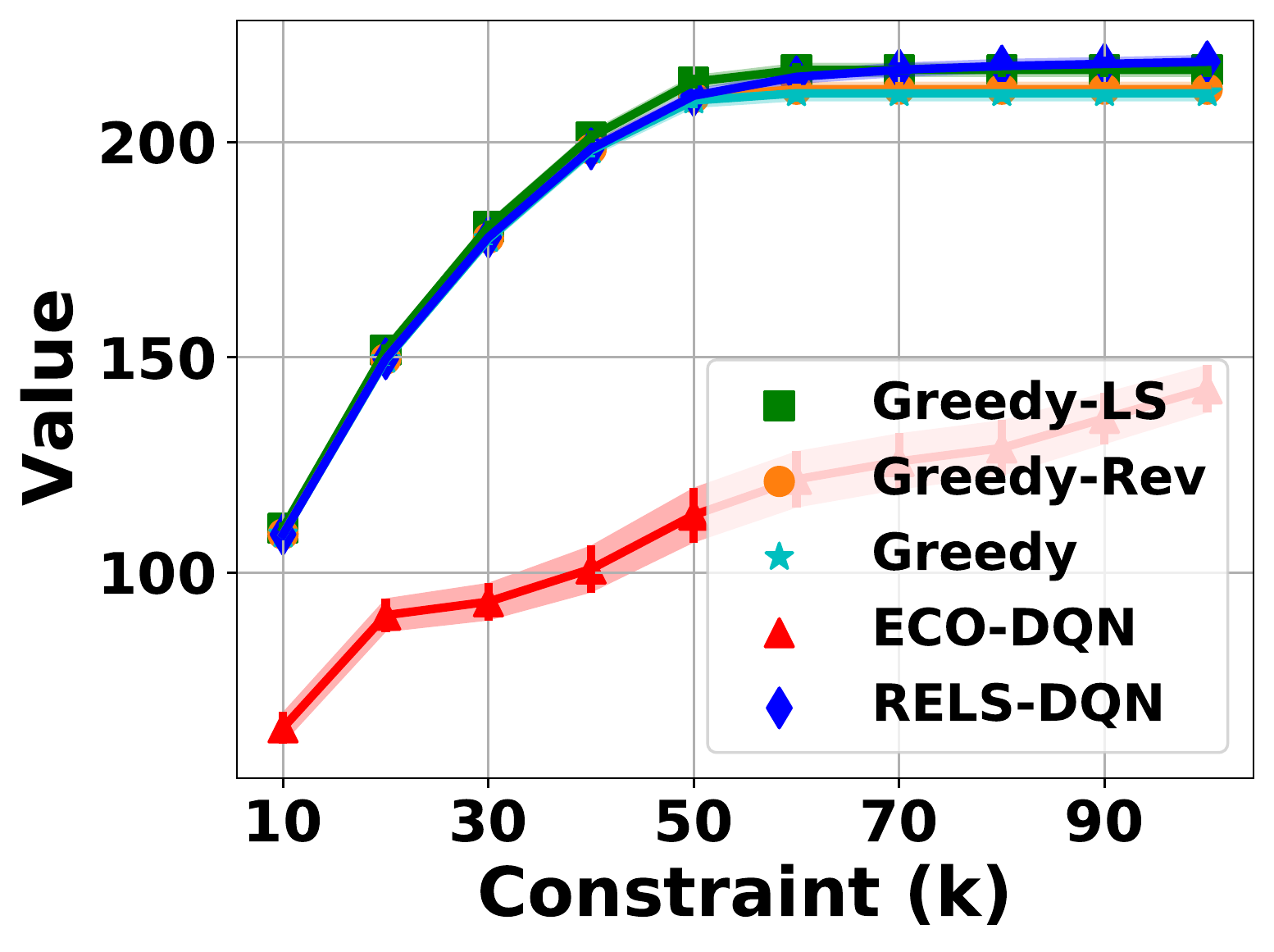}}
    \hfill   
  \subfloat[InfExp (ER200) \label{fig_uncon_infexp2}]{%
        \includegraphics[width=0.24\linewidth]{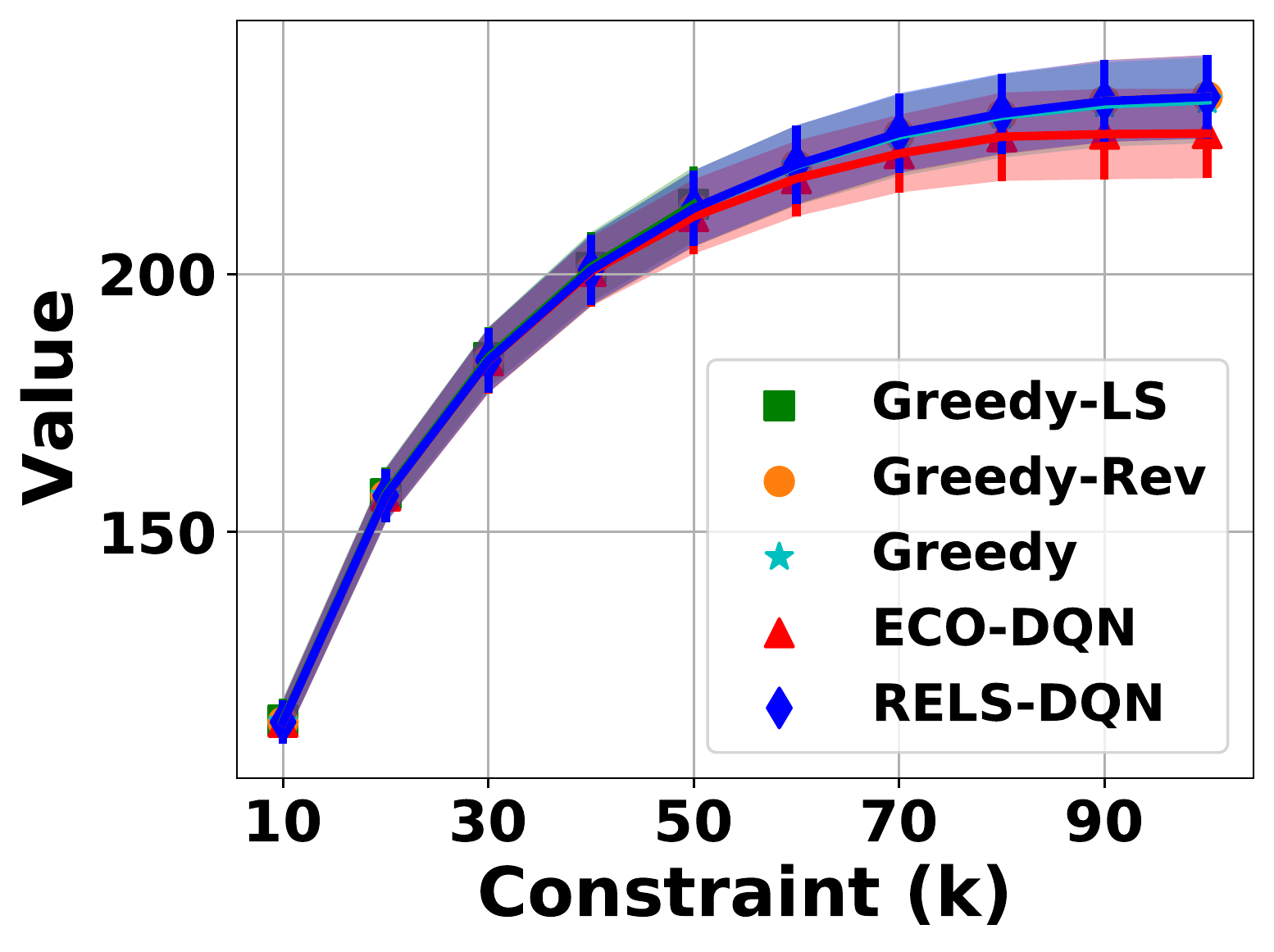}}
        \hfill
    \\
    \subfloat[MaxCut (fb4k)\label{fig_con_maxcut}]{%
       \includegraphics[width=0.24\linewidth]{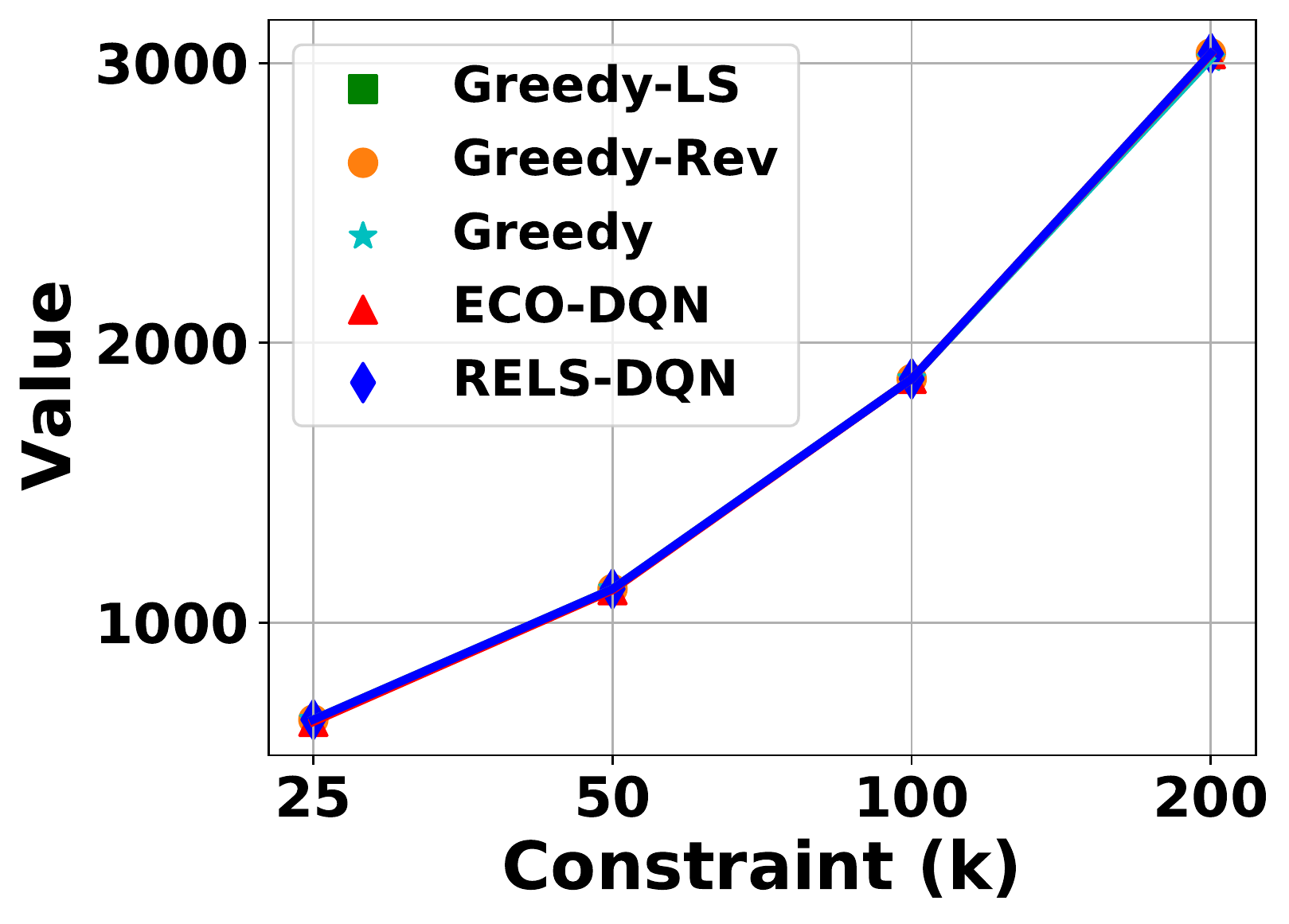}}
    \hfill
  \subfloat[MaxCov (eu1k)\label{fig_con_maxcov}]{%
        \includegraphics[width=0.24\linewidth]{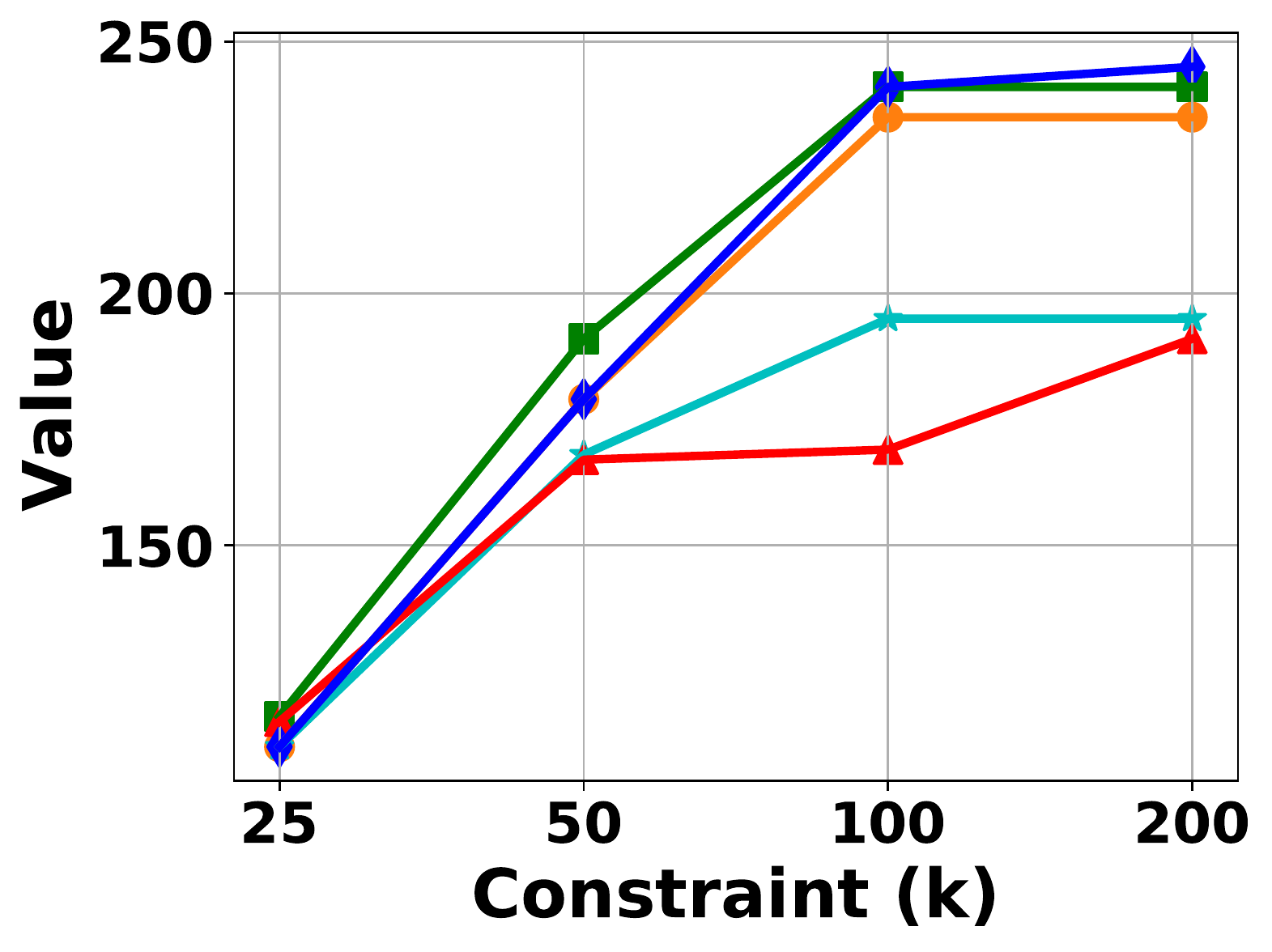}}
    \hfill   
  \subfloat[MovRec (movielen1.7k)\label{fig_con_movmax}]{%
        \includegraphics[width=0.24\linewidth]{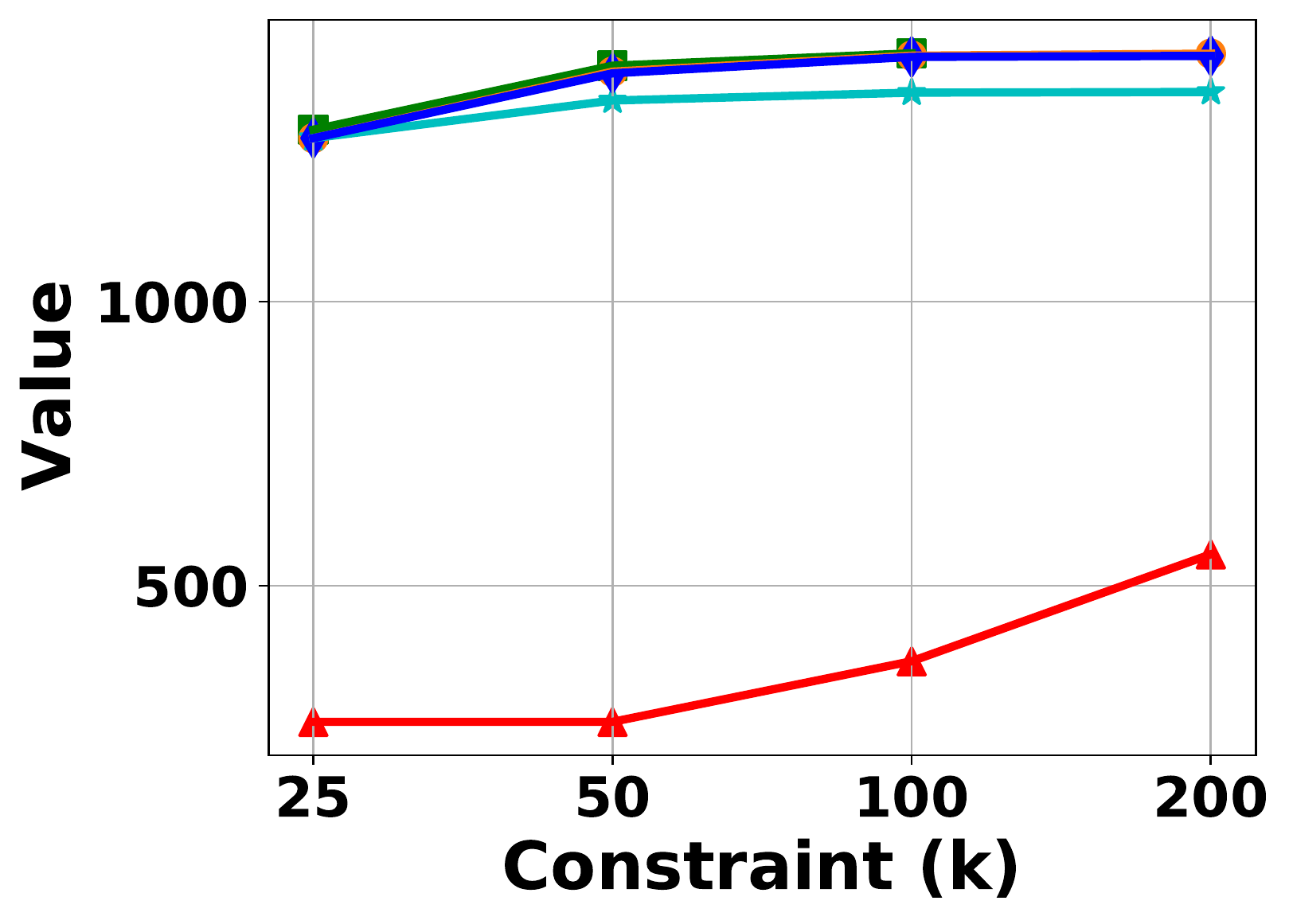}}
    \hfill   
  \subfloat[InfExp (ba300)\label{fig_con_infexp}]{%
        \includegraphics[width=0.24\linewidth]{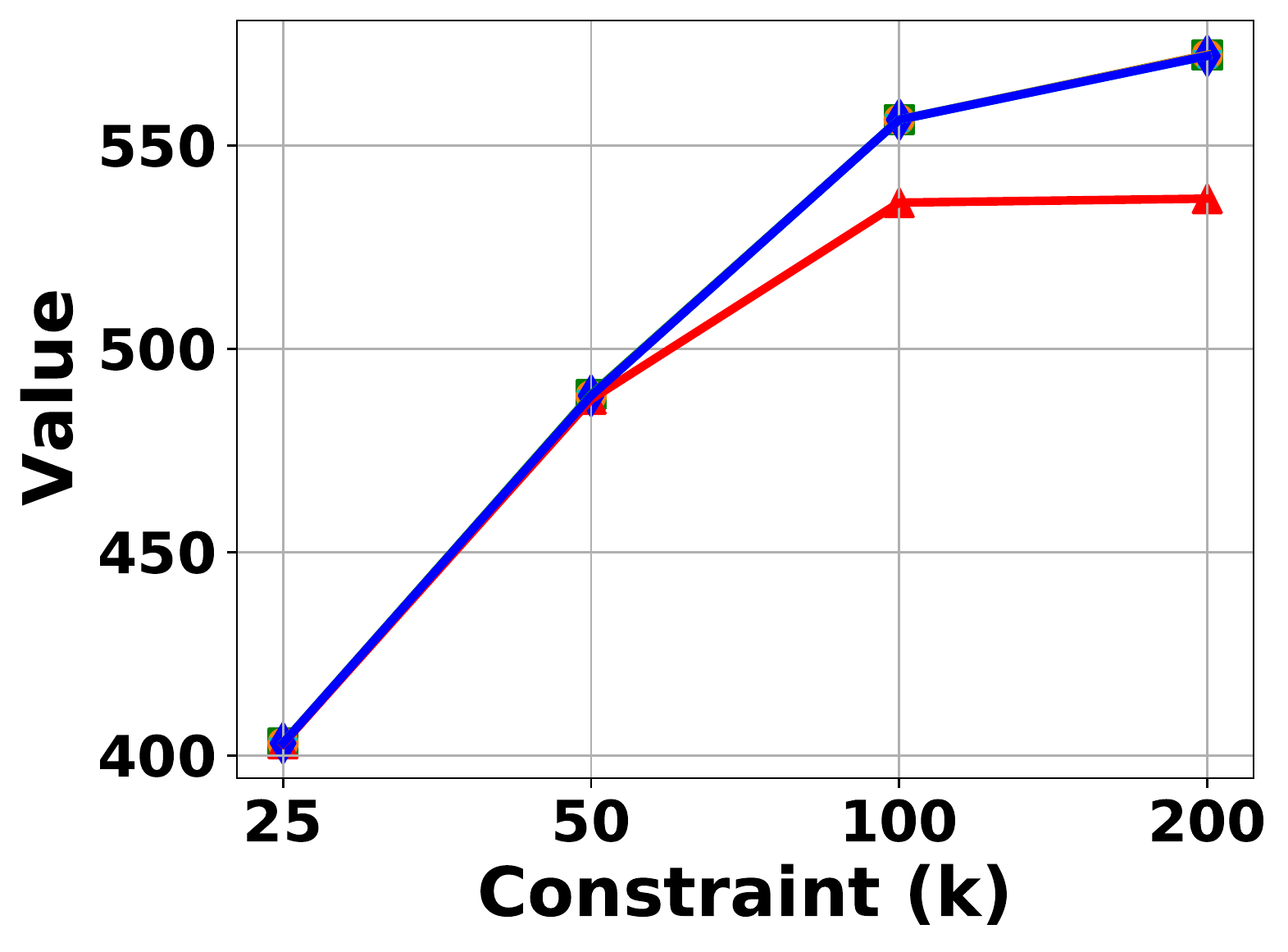}}
  \label{fig-small-con} 
  \vspace{-1.5em}
\end{figure*}

\begin{figure*}[t]
    \centering
    \caption{The runtime results of \lssnn, \eco and the best local search baseline \grls on the large datasets (see Table \ref{tab:dataset}) of four applications. The $timeout$ for each application ($k=\{25, 50, 100, 200\}$) is set to \textbf{24 hours}.}
    \subfloat[MaxCut (fb4k)\label{fig_con_maxcutt}]{%
       \includegraphics[width=0.21\linewidth]{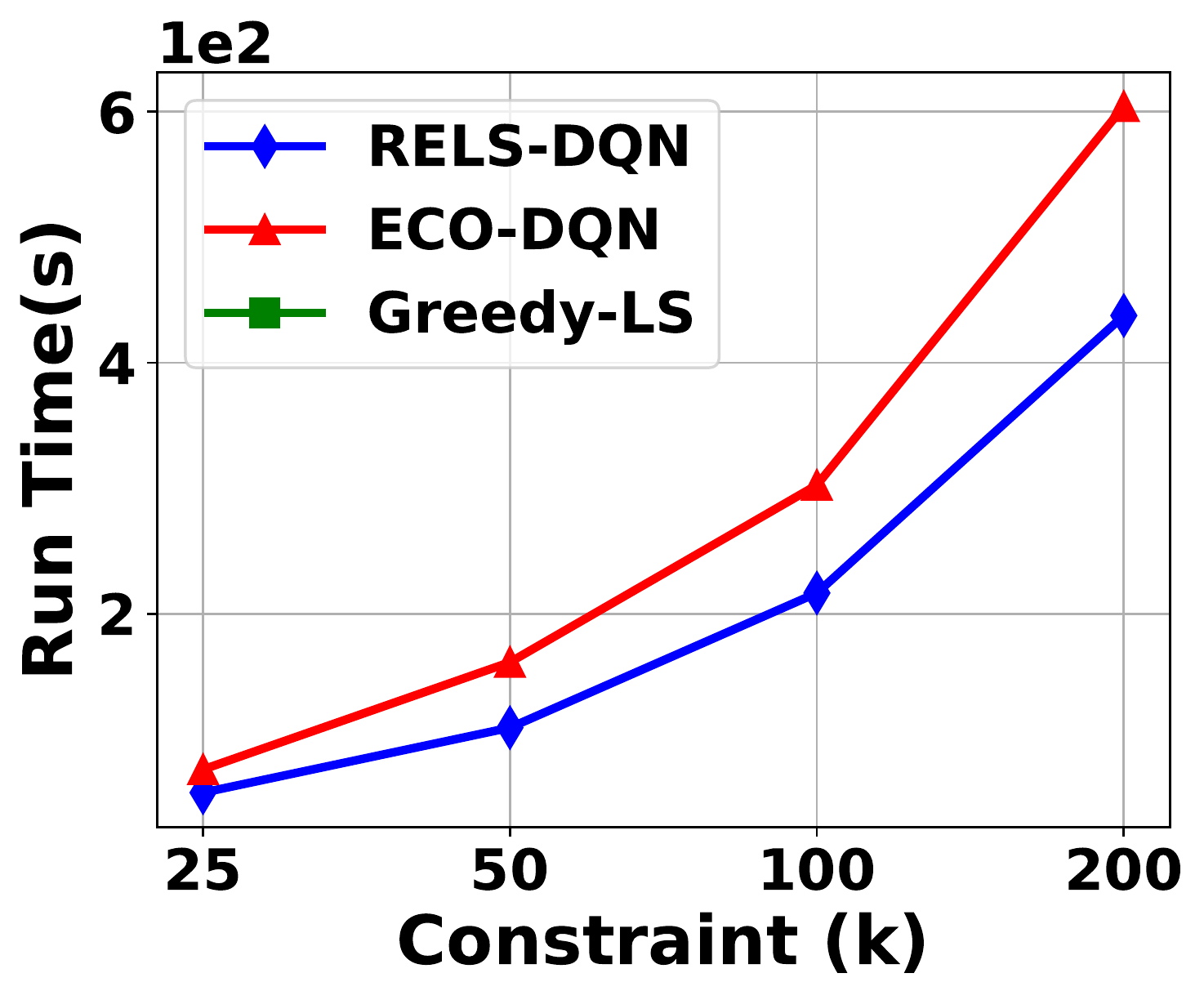}}
    \hfill
  \subfloat[MaxCov (eu1k)\label{fig_con_maxcovt}]{%
        \includegraphics[width=0.24\linewidth]{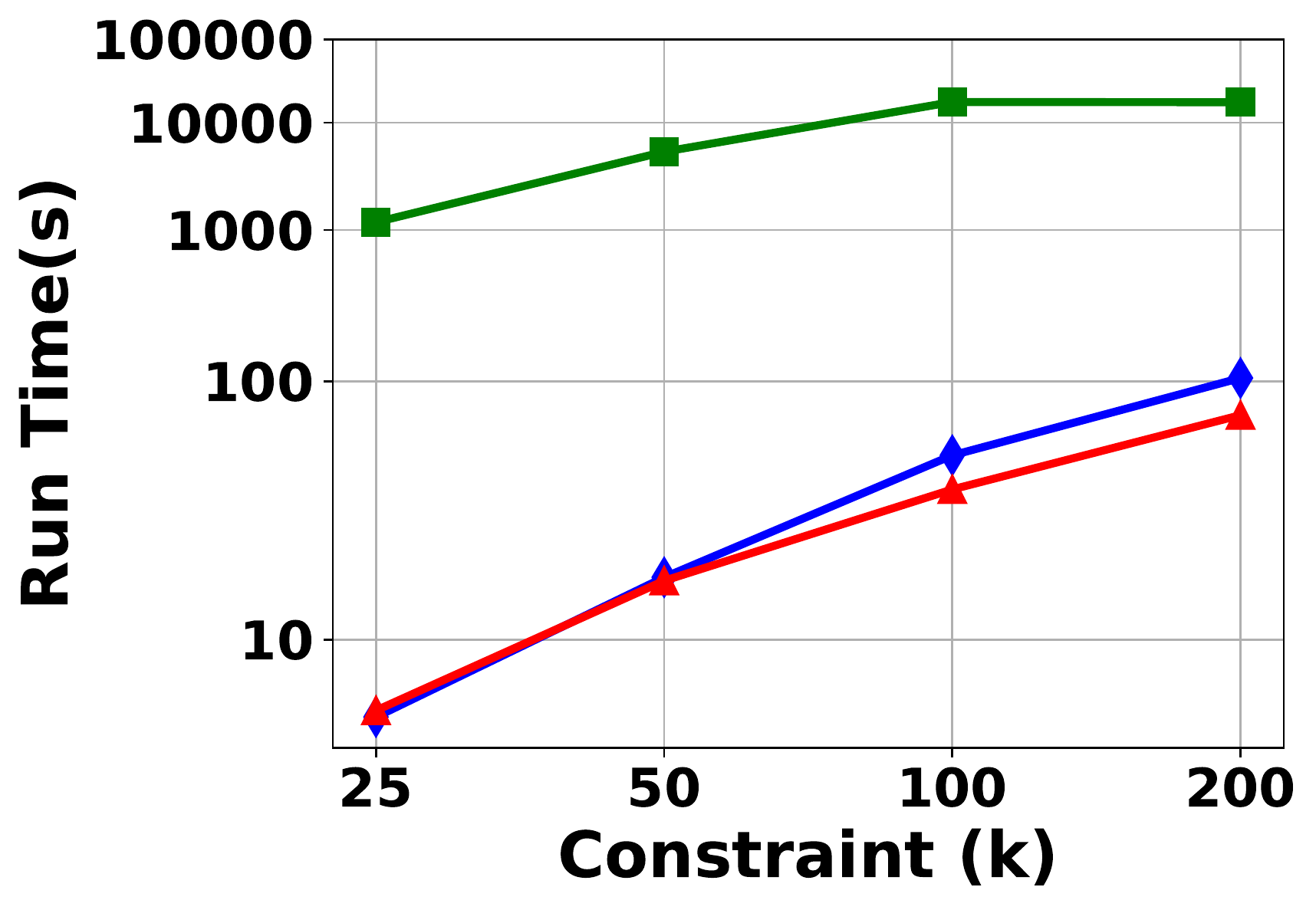}}
    \hfill   
  \subfloat[MovRec (movielen1.7k)\label{fig_con_movmaxt}]{%
        \includegraphics[width=0.24\linewidth]{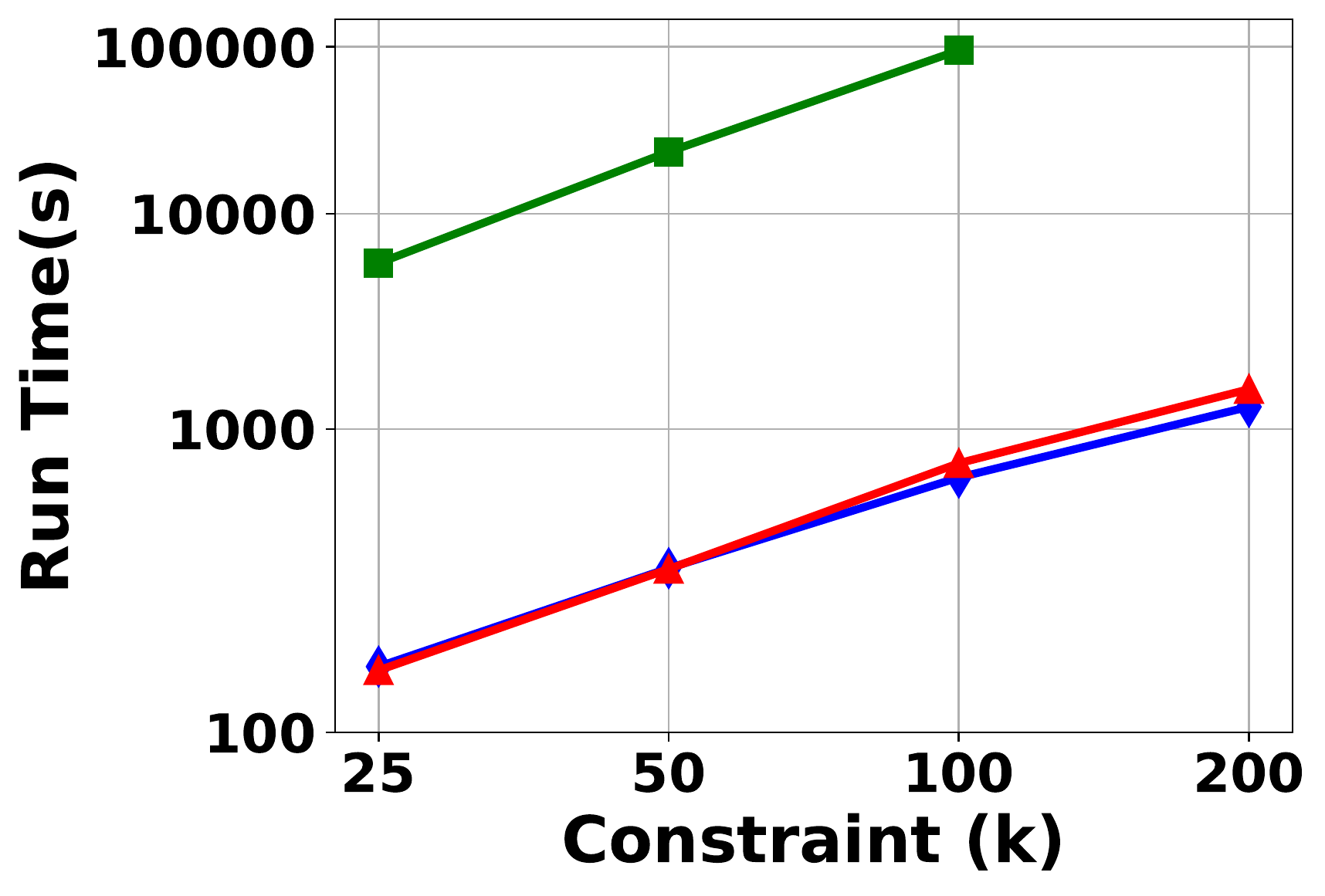}}
    \hfill   
  \subfloat[InfExp (ba300)\label{fig_con_infexpt}]{%
        \includegraphics[width=0.24\linewidth]{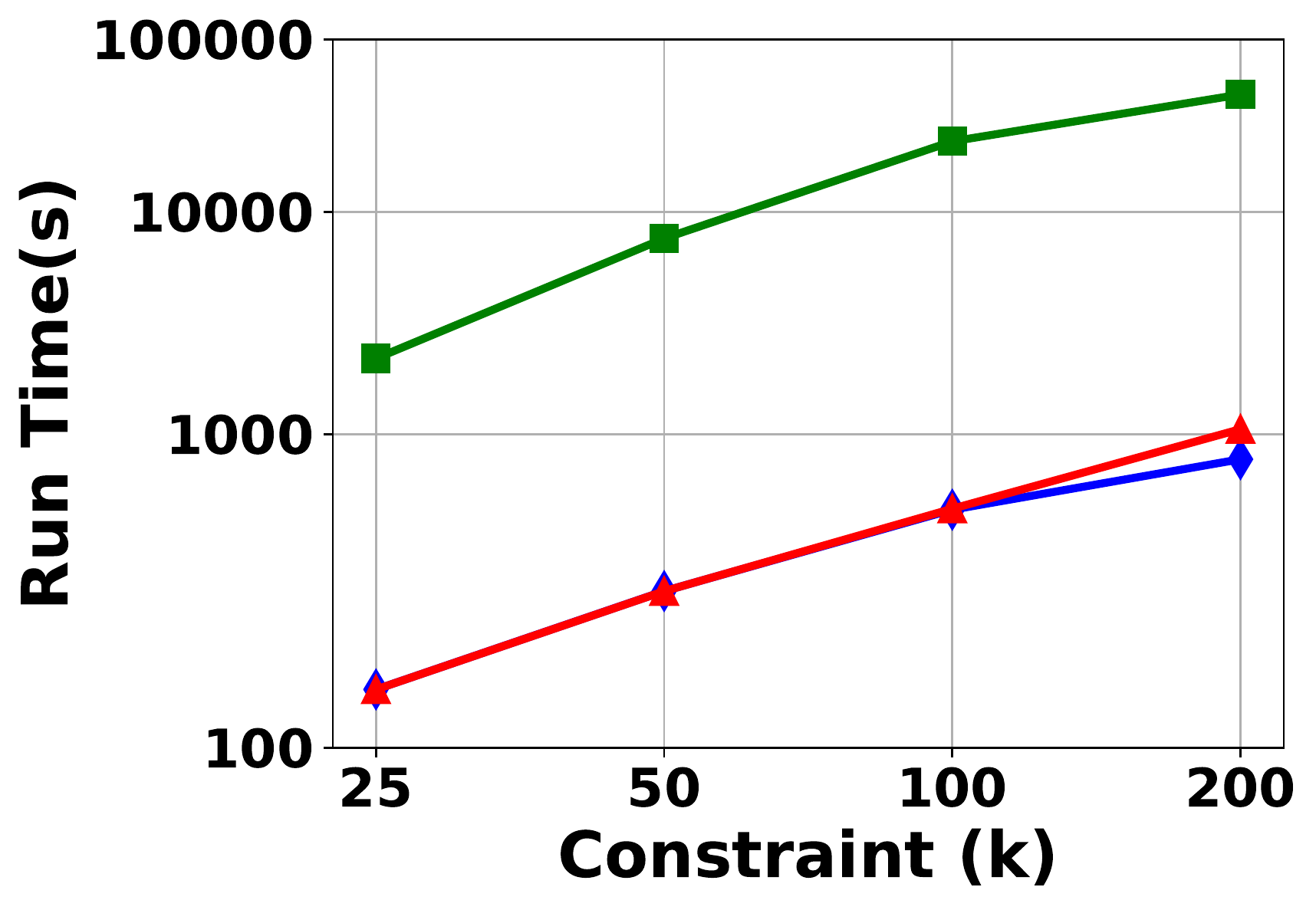}}
  \label{fig-con} 
  \vspace{-1.5em}
\end{figure*}

\subsection{Generalization Evaluation}
\label{conexp}
In this section, we test the ability of \lssnn to provide feasible solutions to various combinatorial problems without application-specific training. To evaluate its generalization, we compare the MaxCut trained \lssnn (ER graph, $n=$40) to the application-specific trained \eco (ER graph, $n=40$), \sg, and \grls and \grev across four applications. We evaluate the models on synthetic and real-world datasets for each application. As part of the synthetic dataset experiments (Figure \ref{fig_uncon_maxcutu}-\ref{fig_uncon_infexp2}), all models are run on a hundred randomly generated synthetic graphs, with the mean solution value plotted.
 
 \textbf{\lssnn Vs. \sg} The empirical comparison of \lssnn to the \sg, the \grev, and the \grls demonstrates our model's ability to provide solutions with values higher than or equal to these general-purpose greedy-based algorithms. As illustrated in Figure \ref{fig-small-con}, due to its lack of local-search capability, \sg is surpassed by \grev and \grls in terms of solution value, with \grls consistently providing similar or outperforming \grev across all applications. For the MaxCut and MaxCov problem on the ER datasets (Figure \ref{fig_uncon_maxcutu}, \ref{fig_uncon_maxcovu}), \lssnn outperforms \grls by 2\% and 3\% respectively on average, with indiscernible results for the other applications.  Additionally, \lssnn provides 5\% and 1.5\% improvement in solution value over \grls for MaxCut and MaxCov on BA datasets, with near-identical results on other applications and large datasets. \lssnn's improvement over \grls can be attributed to its enforced exploration (Alg. \ref{algo:RLS}, Line \ref{line:exp0} - \ref{line:exp1}), which allows the model to search for better solutions even when local-search criteria (Alg. \ref{algo:RLS}, Line \ref{line:exp2} - \ref{line:exp3}) indicate no possibility of adding or removing elements. Across all instances, \lssnn outperforms \sg, \grev, and \grls by 5\%, 2\%, and 1\%, respectively. It is evident from the comparison that \lssnn's local-search capability can provide feasible solutions in various applications without requiring application-specific training, proving its utility as a general-purpose algorithm. Individual application results are provided in Table \ref{tab:summary}.
 
 \textbf{\lssnn Vs. \eco} Across all datasets, \eco and \lssnn provide near-identical solution values for the MaxCut problem. As shown in Figure~\ref{fig_uncon_maxcovu}-\ref{fig_uncon_infexpu}, for MaxCov, MovRec, and InfExp on ER graphs, \lssnn consistently outperforms \eco with a 5\% average improvement in solution value across these instances.  Extending the experiment set to the BA datasets (Figure \ref{fig_uncon_maxcov2} - \ref{fig_uncon_infexp2}), we observe that \lssnn provides an average improvement of 28\% across the MaxCov, MovRec and InfExp application, with this improvement increasing to 114\% for real-world datasets (Figure \ref{fig_con_maxcov} - \ref{fig_con_infexp}). In addition, Table \ref{tab:summary} indicates that the difference between \eco and \lssnn is minimal for ER datasets as the models are trained on ER graphs for each application, and the performance gap increases with changes in graph type and size. The comparison validates the generalization consistency of \lssnn to maintain or outperform application-specific \eco models' performance across a wide range of combinatorial applications and datasets.

\subsection{Efficiency Evaluation}
\label{efficiency}
\textbf{Runtime Evaluation:} We compare the runtime performance of \lssnn to \eco and the best local-search baseline \grls for the real-world datasets. Figure \ref{fig_con_maxcovt} - \ref{fig_con_infexpt} illustrates that \lssnn provides a 264, 64, and 36 times speedup over \grls for MaxCov, MovRec and InfExp respectively. Additionally, for the MaxCut problem, \grls does not complete a single constraint instance within the 24 hours timeout. As shown in Figure \ref{fig_con_maxcutt}, \ref{fig_con_movmaxt} and \ref{fig_con_infexpt}, \lssnn is 40\%, 7\% and 8\% faster than \eco for the MaxCut, MovRec and InfExp problem. Across all applications, \lssnn provides a 10\% runtime improvement over \eco on average. The results exemplify \lssnn's ability to maintain or outperform the \grls in terms of solution value while providing an order-of-magnitude speedup in completion time.

\textbf{Memory evaluation:} We observe the GPU memory usage using the $pynvml$ package. For \eco, we monitor two variants, \eco, which uses an adjacency matrix for its graph representation, and \ecosp, a sparse matrix implementation for MPNN. Table~\ref{tab:gpu} presents the memory usage of these models with ``-'' representing scenarios when an out-of-memory crash occurred. Evaluations are performed on two real-world datasets, fb4k ($n=$4,039 ) and fb22k ($n=$22,470) with batch sizes 1 (Postfix ``-batch1'') and 5. \eco with dense representation (adjacency matrix) runs out of GPU memory on 3 out of 4 instances, with the only instance it completes being fb4k for batch size = 1. While \ecosp drastically improves the memory overhead of \eco, \lssnn on average across all scenarios provides a 30\% improvement over \ecosp.

\begin{table}[h]
    \centering
    \caption{GPU usage (MB).}\label{tab:gpu}
    \begin{tabular}{l p{10mm} p{8mm} p{10mm} p{8mm}}
      \hline\hline 
      Network type & fb4k-batch1 & fb4k & fb22k-batch1 & fb22k \\ [0.5ex] 
      \hline 
      \lssnn & \textbf{1459} & \textbf{1725} & \textbf{3335} & \textbf{11197} \\
      \eco & 10483 & - & - & -\\
      \ecosp & 1571 & 2489 & 5249 & 20663 \\
      \hline 
    \end{tabular}
\end{table}

\section{Conclusion}
\label{conclusion}
In this paper, we have introduced \lssnn, a novel DQN model that demonstrates the ability to perform the local search across a diverse set of combinatorial problems without application-specific training. Unlike existing models, utilizing a compressed feature space and a simpler network architecture allows our model to behave like a general-purpose algorithm for combinatorial problems. Our claims are exemplified through an extensive evaluation of a diverse set of applications that illustrates \lssnn's ability to provide feasible solutions on all four instances of constrained CO problems, with significant speedup over \eco and \grls. In addition, the reduction in memory footprint compared to MPNN models encourage \lssnn's scalability to real-world combinatorial problems. The combined improvement in both generalization and efficiency suggests that the environment states contain valuable information that should not be used as messages in GNN. The future study direction should consider feature-based and structural information separately and analyze their individual effects before exploring their combined impact. On the other hand, our framework also has several limitations. The CO problems that require the ordered sequence are not feasible, such as TSP, because our framework focuses on selecting a subset in an arbitrary order that maximizes the objective function. Another limitation is that the environment representation is element-wise parameters and requires margin gain calculation, which is computationally expensive and the efficiency can be further improved by reducing the number of query times in margin gain calculation.

\clearpage

\bibliographystyle{unsrt}
\bibliography{reference}
\end{document}